\pdfoutput=1 

\documentclass[journal, 12pt, onecolumn]{IEEEtran}
\usepackage{setspace}
\usepackage{booktabs}
\usepackage{arydshln} 
\usepackage{soul}
\ifCLASSINFOpdf
\else
\fi
\usepackage[strings]{underscore}
\usepackage{cite}
\usepackage{amsmath,amssymb}
\usepackage{graphicx}
\usepackage{algorithm}
\usepackage{algorithmicx}
\usepackage[noend]{algpseudocode}
\usepackage{caption}
\usepackage{subcaption}
\usepackage{color}
\usepackage{cleveref}
\usepackage{multicol}
\usepackage{epstopdf}

\usepackage{mathtools}
\usepackage[mathlines]{lineno} 
\newcommand*\patchAmsMathEnvironmentForLineno[1]{%
  \expandafter\let\csname old#1\expandafter\endcsname\csname #1\endcsname
  \expandafter\let\csname oldend#1\expandafter\endcsname\csname end#1\endcsname
  \renewenvironment{#1}%
     {\linenomath\csname old#1\endcsname}%
     {\csname oldend#1\endcsname\endlinenomath}}%
\newcommand*\patchBothAmsMathEnvironmentsForLineno[1]{%
  \patchAmsMathEnvironmentForLineno{#1}%
  \patchAmsMathEnvironmentForLineno{#1*}}%
\AtBeginDocument{%
\patchBothAmsMathEnvironmentsForLineno{equation}%
\patchBothAmsMathEnvironmentsForLineno{align}%
\patchBothAmsMathEnvironmentsForLineno{flalign}%
\patchBothAmsMathEnvironmentsForLineno{alignat}%
\patchBothAmsMathEnvironmentsForLineno{gather}%
\patchBothAmsMathEnvironmentsForLineno{multline}%
}

\begin{document}
\setstcolor{black}

%
\title{Long-Term Inertial Navigation Aided by Dynamics of Flow Field Features}
%
%
%

\author{Zhuoyuan Song,~\IEEEmembership{Student~Member,~IEEE,}
        Kamran Mohseni,~\IEEEmembership{Senior~Member,~IEEE}
\thanks{Z. Song is a graduate student in the Department of Mechanical and Aerospace Engineering, University of Florida, Gainesville, FL, 32611 USA e-mail: (nick.songzy@ufl.edu).}
\thanks{K. Mohseni is the William P. Bushnell Endowed Chaired Professor in the Department of Mechanical and Aerospace Engineering and the Department of
Electrical and Computer Engineering. He is the director of the Institute
for Networked Autonomous Systems, University of Florida, Gainesville, FL, 32611 USA e-mail: (mohseni@ufl.edu).}}

%
%

\markboth{Accepted for publication in \textit{IEEE Journal of Oceanic Engineering}}%
{}

%



\maketitle

\begin{abstract}
A current-aided inertial navigation framework is proposed for small autonomous underwater vehicles in long-duration operations ($>1$ hour), where neither frequent surfacing nor consistent bottom-tracking are available. We instantiate this concept through mid-depth, underwater navigation.  This strategy mitigates dead-reckoning uncertainty of a traditional inertial navigation system by comparing the estimate of local, ambient flow velocity with preloaded ocean current maps. The proposed navigation system is implemented through a marginalized particle filter where the vehicle's states are sequentially tracked along with sensor bias and local turbulence that is not resolved by general flow prediction. The performance of the proposed approach is first analyzed through Monte Carlo simulations in two artificial background flow fields, resembling real-world ocean circulation patterns, superposed with smaller-scale, turbulent components with Kolmogorov energy spectrum. The current-aided navigation scheme significantly improves the dead-reckoning performance of the vehicle even when unresolved, small-scale flow perturbations are present. For a 6-hour navigation with an automotive-grade inertial navigation system, the current-aided navigation scheme results in positioning estimates with under 3\% uncertainty per distance traveled (UDT) in a turbulent, double-gyre flow field, and under 7.3\% UDT in a turbulent, meandering jet flow field.  Further evaluation with field test data and actual ocean simulation analysis demonstrates consistent performance for a 6-hour mission, positioning result with under 25\% UDT for a 24-hour navigation when provided direct heading measurements, and terminal positioning estimate with 16\% UDT at the cost of increased uncertainty at an early stage of the navigation.
\end{abstract}


%
\IEEEpeerreviewmaketitle

\section{Introduction}
Accurate navigation is a prerequisite for mobile robots in order to accomplish basic autonomous tasks such as trajectory tracking and path planning. It is notoriously challenging for an autonomous underwater vehicle (AUV) to maintain a consistently reliable navigation performance~\cite{Chandrasekhar:06a,KinseyJ:06a,TanHP:11a}. The fast attenuation of electromagnetic waves in water creates a poorly illuminated ocean environment while acoustic communication, the primary communication method for most underwater applications, suffers from large delays and high error rates. A universal underwater localization system, similar to the Global Positioning System (GPS), would have been prohibitively expensive to implement with severely limited coverage. 

Existing navigation techniques for AUVs are typically built upon inertial navigation systems (INSs) that usually consist of combinations of accelerometers, gyroscopes, and magnetometers. Sampling noise of these sensors is propagated and amplified during the integration steps of dead-reckoning. Thus, an INS usually suffers from severe drift error if no reliable external references are available. Intermittent corrections of the state estimation error with exteroceptive sensors are therefore necessary for maintaining a reasonably accurate navigation performance. To this end, several techniques have been proposed~\cite{PaullL:14a}, including sonar-based methods~\cite{DonovanGT:12a,Newman:05a}, and surface-agent-aided methods~\cite{Bahr:09a,FallonMF:10a,ArrichielloF:12a}. However, one common limitation of these techniques is that the vehicle must stay close to either artificial beacons (e.g.~long baseline), static landmarks (e.g.~underwater simultaneous localization and mapping (SLAM)), or the sea bottom (e.g.~Doppler velocity log (DVL) bottom-tracking). 

When a mobile robot utilizes external features to mitigate its navigation error, an aided navigation system is established. Depending on the characteristics of the features, the resultant aided navigation systems are different. The vast majority of existing aided navigation schemes can be classified as the first category, which utilizes the static properties of navigation features (e.g. positions of landmarks~\cite{DellaertF:99a,Durrant-WhyteH:06a,ClausB:15a}). Nevertheless, there usually are situations where the robot does not encounter a sufficient number of, or any, such features for the navigation system to reference, hence static-feature-based navigation schemes become inapplicable. The second category of aided navigation systems utilizes features' states that are time-dependent, such as the location of moving beacons~\cite{KussatNH:05a,FallonMF:10a}, or neighboring robots~\cite{MourikisAI:06a,WymeerschH:09a}. However, the dynamics governing the changes in states of these features are often not utilized. This motivates us to explore a new category of aided navigation systems that exploits the dynamics of environmental features in robot navigation problems where the environment and the features therein are considered as another dynamic system interacting with the robot.

One of the key factors that affects the performance of an aided navigation method is the frequency at which a robot has access to the features for reference. Navigation methods based on physical landmarks limit the robot's navigation range to the milieu of these features. In contrast, methods based on field features utilize a certain type of background field, either natural or artificial, as the navigation reference. The robot is able to perform correction for positioning error at any location by measuring the local strength of the background field~\cite{ChungJ:11,Gutmann:12b}. This type of methods significantly expands the robot's navigation range with a prerequisite that a map of the field needs to be known beforehand. When the background field is spatiotemporally changing, it is necessary to understand the dynamics of the field in order to extract useful features for navigation.

Traditionally, omnipresent ocean currents are usually considered as a major culprit behind the performance degradation in control and navigation of small underwater vehicles. Due to the lack of understanding of ocean current dynamics, background flows were usually treated as small disturbances for conventional underwater vehicles. Since the emergence of smaller-size AUVs, the impact of ocean currents on a vehicle's motion has drawn certain amount of attention because the velocities of ocean currents sometimes exceed the maximum speed of these actuation-limited AUVs \cite{FossenTI:91a,FossenTI:11a}. Fortunately, the ever-increasing computational power and the emergence of multiple in-situ ocean-sampling techniques have contributed to the recent progress in ocean general circulation models (OGCMs) that simulate the general movement of large-scale ocean currents. By closing this knowledge gap, researchers have been utilizing the background flow dynamics in various aspects of underwater robotics, including path planning and vehicle guidance~\cite{InancT:05a,Mohseni:13r,LermusiauxPFJ:16a}, multi-vehicle cooperative control~\cite{PaleyD:07a,Mohseni:10t,Song:17a}, optimal sensing~\cite{FiorelliE:06a,LeonardNE:07a,ZhangF:07a,PaleyDA:08a}, sensor network mobility analysis~\cite{CarusoA:08a,Song:15a}, and mobile sensor allocation~\cite{Mallory:13a}. Nonetheless, to the best of our knowledge, the background flow dynamics has not been well exploited for underwater localization and navigation problems with only a few exceptions~\cite{HegrenaesO:09a,StanwayMJ:12a,ChangD:15a,MedagodaL:15a,Medagoda:16a}.

\begin{figure}
\centering
\includegraphics[trim={0, 0, 0, 10mm}, clip, width = 0.5\linewidth]{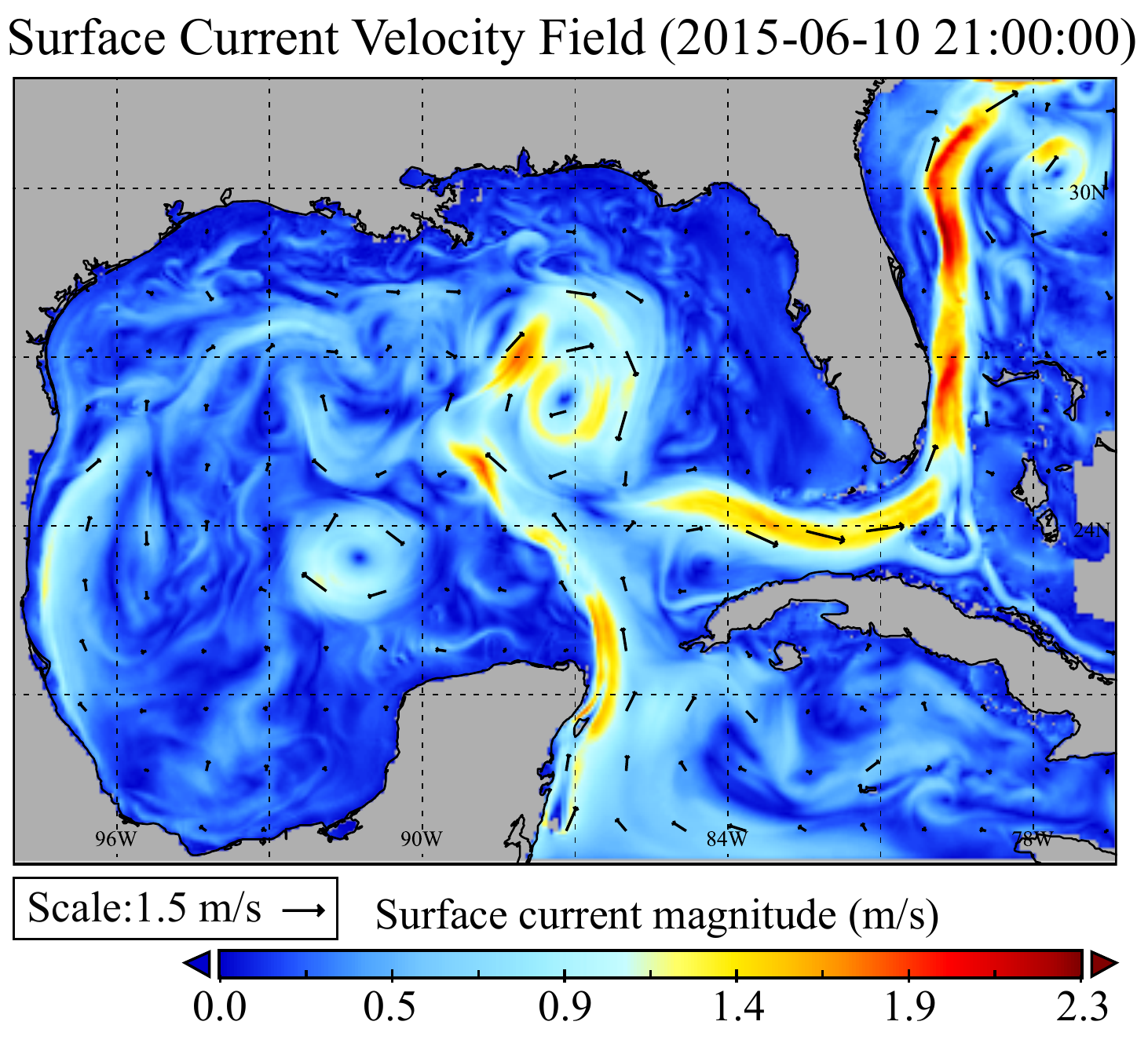}
\caption{Velocity field of the linearly-interpolated surface current of the Gulf of Mexico at 21:00:00 on June 10, 2015 predicted by HYCOM Gulf of Mexico.} 
\label{fig:forecast}
\end{figure}

OGCMs simulate large-scale ocean movement by numerically solving the Navier-Stokes equations with multiple boundary conditions, and by assimilating real-time, in-situ measurement data obtained by satellites, surface vessels, underwater gliders, etc. Particularly, Lagrangian instruments, such as drifter buoys (e.g. the Global Drifter Program by NOAA AOML Physical Oceanographic Division) or floats, are usually used as proxies for ocean flow advection and have contributed significantly to the study of general ocean circulations \cite{Swallow:55a,PattersonSL:85a,DavisRE:91a,LavenderKL:00a,VenezianiM:04a}. Nowadays, publicly available OGCMs can provide forecasts of global ocean currents for up to six days with an average spatial resolution of 3 kilometers and a temporal resolution of 3 hours or less \cite{ChassignetEP:09a,MehraA:10a}. With smaller-scale regional models, higher spatiotemporal resolution often can be achieved. Fig.~\ref{fig:forecast} shows the velocity field of the linearly-interpolated surface currents of the Gulf of Mexico at 9 p.m. on June 10th, 2015 predicted by HYCOM~\cite{ChassignetEP:09a}. 

During long-range AUV operations, such as long-duration underwater sampling or surveillance, it is often the case that frequent surfacing for dead-reckoning error correction with GPS is not desirable, the limited number of structural features prevents the implementation of underwater SLAM, and consistent DVL bottom-tracking is not always available, if at all. In situations like these, the ubiquitous background flow may be utilized to mitigate the dead-reckoning error when the general ocean circulation is predictable. 

\begin{figure}
\centering
\includegraphics[width=0.4\linewidth]{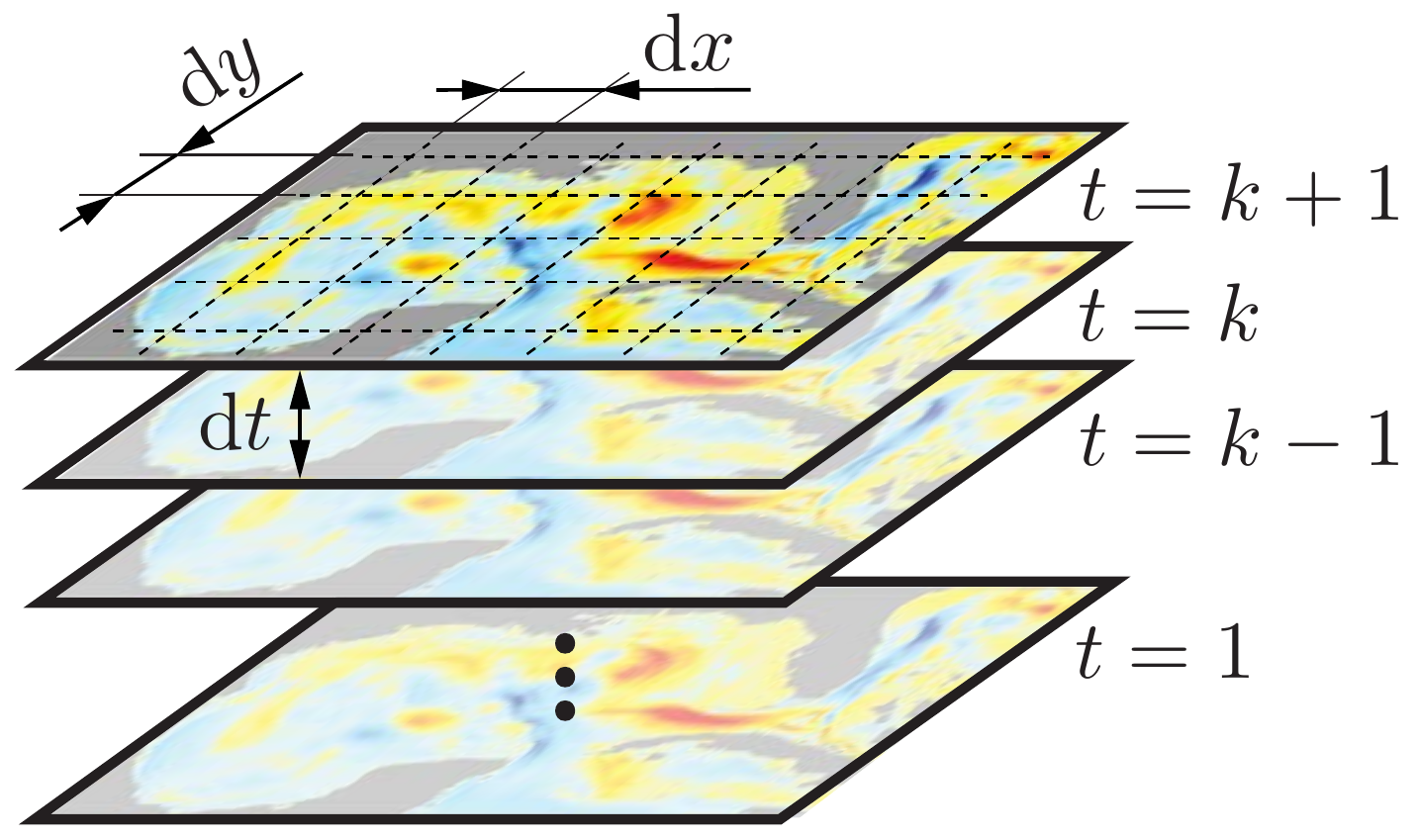}
\vspace{-0.0in}
\caption{A conceptual illustration of a time-tagged current velocity map series generated an OGCM. Variables $\mathrm{d}x$ and $\mathrm{d}y$ depend on the longitudinal and latitudinal resolution of the ocean model, and $\mathrm{d}t$ depends on the temporal resolution.}
\label{fig:mapgrid}
\end{figure}

In this work, we demonstrate the concept of current-aided inertial navigation.  This system improves the inertial navigation performance by utilizing local measurements of relative flow velocities and predictions of the flow field provided by OGCMs~\cite{Song:14b,Song:14c}. Before the deployment of an AUV, a sequence of current velocity maps of the navigation domain can be predicted by an OGCM, such as RTOFS-Atlanta~\cite{MehraA:10a} or HYCOM~\cite{ChassignetEP:09a}; (see Fig.~\ref{fig:mapgrid}). During operations, the AUV estimates local, absolute flow velocities through on-board sensors and deduces its own states from preloaded current velocity maps. A marginalized sequential Monte Carlo method is adopted to implement such a recursive Bayesian estimator. 

Different from this work, Chang et al.~\cite{ChangD:15a} proposed a real-time guidance system for underwater gliders using predictive ocean models in combination with glider-derived flow estimate to improve the gliders' navigation performance in complex flow regions with large spatial and temporal flow changes. Their navigation scheme focuses on gliders that perform dead reckoning using depth and attitude sensors instead of initial navigation systems and requires the vehicles to resurface at regular intervals. Medagoda et al.~\cite{StanwayMJ:12a,Medagoda:16a} focused on the navigation performance during vehicle descent by utilizing the fact that ocean currents are approximately invariant within a short period of time.  Similarly, Hegren{\ae}s and Berglund~\cite{HegrenaesO:09a} only considered their DVL water-tracking technique as an intermediate approach to bridge the gap between other more reliable global localization techniques involving GPS or USBL.  By including the knowledge of ocean general circulation, we aim to provide long-term performance improvement over inertial navigation systems alone. We consider our approach and the aforementioned existing methods mutually complementary. Major contributions of our work include a generalized inertial navigation system based on ocean general circulation, treatment of unpredictable turbulence, and a systematic performance analysis of a current-aided navigation estimator in a simulated environment.

In what follows, the current-aided navigation system is derived based on a recursive Bayesian structure in Section~\ref{sec:formulation}. A marginalized particle filter (MPF) realization is presented and the parametric Cram\'{e}r-Rao lower bound (CRLB) is derived for a reduced system. Section~\ref{sec:simulation} includes a systematical performance evaluation of the current-aided inertial navigation system in two turbulent flow fields with pre-generated sensor samples. The proposed current-aided MPF is compared to an extended Kalman filter (EKF) based method and the parametric CRLB.  Section~\ref{sec:experiment} further justifies the feasibility of the current-aided navigation scheme through a simulated experiment with field test data and OGCM results.  Finally, we conclude the paper in Section~\ref{sec:conclusion}.

\section{Current-Aided Inertial Navigation}\label{sec:formulation}
Current-aided inertial navigation can be formulated as a partially observable, nonlinear state estimation problem. We first mathematically formulate it in probability theory. Targeting at estimating a probability distribution of the vehicle's states given proprioceptive and exteroceptive sensor measurements, a recursive Bayesian structure is adopted to guide the entire state estimation process. Sensor bias and unmodeled local turbulence are tracked in an online fashion along with the vehicle's states. This general formulation will then be realized through numerical approximation since an optimal solution to the target Bayesian recursion in not easily tractable. 

\subsection{Preliminaries on Probabilistic Formulation}
\begin{figure}
\centering
\includegraphics[width = 0.5\linewidth]{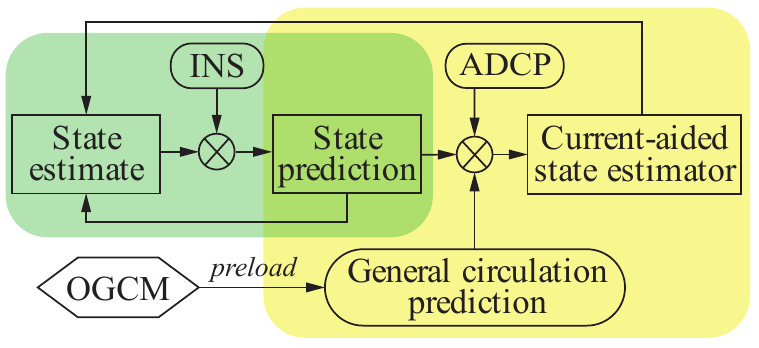}
\caption{Illustration of the concept of current-aided inertial navigation.}
\label{fig:flowChart}
\end{figure}

Typically, a mid- to long-range AUV is equipped with an INS that provides acceleration and angular velocity measurements for purposes including state estimation and disturbance rejection. Magnetometers and pressure sensors are also often used to provide direct attitude and depth estimates, respectively. We assume that the vehicle can measure the relative flow velocities of the ambient fluid with respect to itself. The locations of these measurements should be at least one body-length away from the vehicle-fluid boundary layer such that the presence of the vehicle does not significantly alter the flow velocities at the locations of measurement. Such measurements can be obtained by using an acoustic Doppler current profiler (ADCP) or the current profiling function on some DVLs, which are nowadays becoming standard for small AUVs. With such a sensor suite, improved navigation performance, although with theoretically unbounded uncertainty, can be potentially achieved in mid-depth applications utilizing methods proposed by Stanway~\cite{StanwayMJ:12a} or Medagoda et al.~\cite{Medagoda:16a} that overlap the consecutive current profiling measurements.  We aim at providing long-term performance improvement to inertial navigation systems by taking advantage of the information contained in large-scale ocean current simulation.  This requires the current forecast maps to be preloaded onto the vehicle, which we consider practical given the current state of data storage capacity for compact devices. This design concept is illustrated in Fig.~\ref{fig:flowChart}.

The vehicle's state vector at time $k\in\mathbb{N}^{+}$ can be generally denoted as $\boldsymbol{x}_k\in \mathbb{R}^n$. This state vector contains the vehicle's position, sensor bias, and local turbulence. We divide them into two subgroups such that $\boldsymbol{x}_k =[\boldsymbol{x}^\text{KF}_k, \boldsymbol{x}^\text{PF}_k]$, where $\boldsymbol{x}^\text{PF}_k$ denotes the location of the vehicle to be estimated using a particle filter (PF), and $\boldsymbol{x}^\text{KF}_k$ contains the remaining states including vehicle velocity, sensor bias, etc.~that can be estimated using parametric state estimators such as a Kalman filter (KF) or its extensions. This is motivated by the fact that the vehicle's location will be used for deducing the local ocean current based on the preloaded forecast maps that are typically grid maps without an analytical description. Although analytical approximation may be obtainable for the digital current maps such that an extended Kalman filter (EKF) or alike can be applied for tracking all of the states, the additional computations and the potential inaccuracy in linearizing the flow map (typically highly nonlinear) may keep practitioners from doing so. All the sensor measurements, including accelerations, angular velocities, and relative current velocities can be denoted by $\boldsymbol{z}_k\in\mathbb{R}^l$.  To simplify the notation, we group the history of vehicle states and measurements up to time step $k$ as ${X}_k = \{\boldsymbol{x}_0, ..., \boldsymbol{x}_k\}$ and ${Z}_k = \{\boldsymbol{z}_1, ..., \boldsymbol{z}_k\}$, respectively. The following derivations are based upon two standard assumptions:
\begin{enumerate}
	\item The change of vehicle states can be modeled as a Markov process such that $\Pr(\boldsymbol{x}_k\,|\,{X}_{k-1},{Z}_{k-1}) = \Pr(\boldsymbol{x}_k\,|\,\boldsymbol{x}_{k-1})$;
	\item Observations at different time steps are mutually independent conditional on the vehicle's states such that $\Pr(\boldsymbol{z}_k\,|\,{Z}_{k-1},{X}_{k}) = \Pr(\boldsymbol{z}_k\,|\,\boldsymbol{x}_k)$.
\end{enumerate}

The current-aided inertial navigation problem can now be formulated to estimate the conditional probability density function (pdf) $p({X}_{k}\,|\,{Z}_{k})$. We take the factorization of the target pdf such that
\begin{equation}\label{eq:factorization}
p({X}_{k}\,|\,{Z}_{k}) = p({X}^\text{KF}_{k}\,|\,{X}^\text{PF}_{k},{Z}_{k}) \cdot p({X}^\text{PF}_{k}\,|\,{Z}_{k}).
\end{equation}
This factorization allows us to marginalize the location of the vehicle such that all the other states can be estimated with a Kalman filter or alike. Based on the aforementioned assumptions, a Bayesian recursion can be found for the second term in (\ref{eq:factorization}) as
\begin{equation}
p({X}^\text{PF}_{k}\,|\,{Z}_{k}) \propto p(\boldsymbol{z}_{k}\,|\,\boldsymbol{x}^\text{PF}_{k}) \cdot p(\boldsymbol{x}^\text{PF}_{k}\,|\,\boldsymbol{x}^\text{PF}_{k-1}) \cdot p({X}^\text{PF}_{k-1}\,|\,{Z}_{k-1}).
\end{equation}
Due to the nature of an ocean current map, the posterior pdf $p({X}^\text{PF}_{k}\,|\,{Z}_{k})$ is almost always multi-modal. A series of random measure of it can be defined as $\{\boldsymbol{x}^{\text{PF},i}_{k}, \, w^i_k\}^{N_p}_{i = 1}$ where $w^i_k$ can be considered as the weights of the corresponding support points (particles), $\boldsymbol{x}^{\text{PF},i}_{k}$, satisfying $\sum^{N_P}_iw^i_k = 1$, where $N_p \in \mathbb{Z}^+$ denotes the number of support points. Thus, a discrete approximation of the posterior pdf $p({X}^\text{PF}_{k}\,|\,{Z}_{k})$ can be obtained as 
\begin{equation}
p({X}^\text{PF}_{k}\,|\,{Z}_{k}) \approx \sum^{N_p}_{i=1} w^i_k \delta_{{X}^{\text{PF},i}_{k}}({X}^{\text{PF}}_{k}).
\end{equation}
A sequential importance sampling (SIS) scheme~\cite{DoucetA:01a} can be applied by defining the particle weights as $w^i_k = p({X}^{\text{PF},i}_{k}\,|\,{Z}_{k})/q({X}^{\text{PF},i}_{k}\,|\,{Z}_{k})$ such that sampling from the proposal distribution $q({X}^{\text{PF},i}_{k}\,|\,{Z}_{k})$ is easy to accomplish. In the case of filtering, where one only cares about the present location of the vehicle, the weights can be updated recursively at the arrival of new observations such that 
\begin{equation}\label{eq:weights}
w^i_{k+1} \propto w^i_k \dfrac{p(\boldsymbol{z}_{k+1}\,|\,\boldsymbol{x}^{\text{PF},i}_{k+1})\,p(\boldsymbol{x}^{\text{PF},i}_{k+1}\,|\,\boldsymbol{x}^{\text{PF},i}_{k})}{q(\boldsymbol{x}^{\text{PF},i}_{k+1}\,|\,\boldsymbol{x}^{\text{PF},i}_{k},\boldsymbol{z}_{k+1})}.
\end{equation}
A proper choice of the proposal distribution has been proven crucial in many studies~\cite{RobertC:2013a}. One widely adopted option in navigation applications is the prior distribution $p(\boldsymbol{x}^{\text{PF},i}_{k+1}\,|\,\boldsymbol{x}^{\text{PF},i}_{k})$, which leads to a simply yet practically efficient update recursion for particle weights $w^i_{k+1} \propto w^i_k \cdot p(\boldsymbol{z}_{k+1}\,|\,\boldsymbol{x}^{\text{PF},i}_{k+1})$. Such a prior distribution has shown to be suitable for sequential estimation problems with observation noise larger than process noise, which is the case for our application. 

Conditional on $p({X}^{\text{PF},i}_{k+1})$, the joint conditional pdf of the remaining states $p({X}^\text{KF}_{k}\,|\,{X}^\text{PF}_{k},{Z}_{k})$ can be sequentially estimated using parametric estimators, and the implementation details will be discussed in Section~\ref{sec:implementation}. It is worth motioning that, factorization~(\ref{eq:factorization}) is different from its conventional form where nonlinear states are marginalized such that all the remaining states are linear and optimal estimators are applicable~\cite{DoucetA:00a}. In our case, including the vehicle's attitude as an estimation state makes the remaining states nonlinear due to coordinate transformations. Tracking the vehicle's attitude using nonparametric methods along with the vehicle's location will lead to larger numbers of sampling dimension (e.g. from 3 to 6 in a three-dimensional case), which requires significantly more computational resources. Instead, we directly tackle the nonlinearity in the remaining states using an EKF. The approximation inaccuracy associated the linearization step can be considered negligible for AUV applications since heading changes are typically small due to large hydrodynamic damping effects. If available, additional heading correction can be utilized to provide bounded attitude estimate, allowing for implementation of Kalman filters for the remaining linear states.

\subsection{System Models}
One major distinction between our current-aided inertial navigation strategy and many existing terrain-based navigation methods is that the navigation references are dynamic targets, i.e.~a spatiotemporally changing flow field. Due to the scarcity of reliable alternative features in mid-depth ocean, we take advantage of our knowledge in the dynamics of the general ocean circulation that has been accumulated for decades. However, the chaotic nature of ocean currents makes it impossible for a numerical model to capture the exact motion of the flows on all scales. This requires deliberation in the utilization of an ocean circulation forecast as the navigation reference.

In fluid dynamics, the velocity of a turbulent flow field can be separated based on Reynolds decomposition as $\boldsymbol{U}(\boldsymbol{x},t) = \overline{\boldsymbol{U}}(\boldsymbol{x}) + \boldsymbol{u}(\boldsymbol{x},t)$, where $\overline{\boldsymbol{U}}$ is the large-scale, slowly-varying steady component and $\boldsymbol{u}$ includes the small-scale, local perturbations. In this study, we assume that the ocean circulation models can provide predictions of the large-scale currents, which the vehicle will utilize as navigation references. The unmodeled turbulent flow component at the vehicle's locations will be estimated online. Ocean current forecast maps can be generated and loaded on to the vehicle before deployment. As we will be showing soon, knowledge about the ocean turbulence statistics in the navigation domain will further benefit the estimation of the unmodeled turbulence.

To simplify the demonstration of the current-aided navigation concept, we focus on two-dimensional, horizontal navigation of an AUV. This does not prevent the extension of the proposed method to three-dimensional cases since the vehicle depth can generally be estimated independently using absolute pressure measurements and the coupled six-degrees-of-freedom motion can be tackled by simply extending this two-dimensional treatment to a higher-dimensional state space. Such an extension is particularly straightforward for AUVs that do not relay on control surfaces (e.g. AUVs designed by our research group~\cite{Mohseni:11e,Song:16a}).

The state vector to be estimated can be defined as $\boldsymbol{x}_k = [\boldsymbol{p}^{\{n\}\top}_k, \boldsymbol{v}^{\{n\}\top}_k, \psi_k, \boldsymbol{b}^{\{b\}\top}_{\boldsymbol{a},k}, b_{r,k},\boldsymbol{b}^{\{b\}\top}_{\boldsymbol{z},k}, \boldsymbol{u}^{\{n\}\top}_{c,k}]^\top \in \mathbb{R}^{12}$. Here, $\boldsymbol{p}^{\{n\}}_k = [x,y]^\top$, $\boldsymbol{v}^{\{n\}}_k = [v_x, v_y]^\top$, and $\psi_k$ represent the location, linear velocity, and heading of the vehicle in the inertial frame $\{n\}$, respectively. $\boldsymbol{b}^{\{b\}}_{\boldsymbol{a},k}$, $b_{r,k}$ and $\boldsymbol{b}^{\{b\}}_{\boldsymbol{z},k}$ are sensor bias of acceleration measurement $\boldsymbol{a}^{\{b\}}_k$, yaw rate measurement $r^{\{b\}}$, and relative current velocity measurement $\boldsymbol{z}^{\{b\}}$ in the body frame $\{b\}$. The last state, $\boldsymbol{u}^{\{n\}}_{c,k} = [u_x, u_y]^\top$, is the local, unmodeled turbulent flow component at the vehicle's location represented in the inertial frame. 

The discrete system dynamics for the vehicle can be expressed as
\begin{align}
\hat{\boldsymbol{p}}^{\{n\},-}_{k+1} &= \hat{\boldsymbol{p}}^{\{n\},+}_{k} + \hat{\boldsymbol{v}}^{\{n\},+}_{k}\delta t, \label{eq:dynamicsStart}\\
\hat{\boldsymbol{v}}^{\{n\},-}_{k+1} &= \hat{\boldsymbol{v}}^{\{n\},+}_{k} + R^n_b(\hat{\psi}^+_{k})(\boldsymbol{a}^{\{b\}}_k - \hat{\boldsymbol{b}}^{\{b\},+}_{\boldsymbol{a},k} - \boldsymbol{w}_{\boldsymbol{a},k})\delta t, \\
\hat{\psi}^-_{k+1} &= \hat{\psi}^+_{k} + (r_k - \hat{b}^+_{r,k} - w_{r,k}) \, \delta t,
\end{align}
where $R^n_b(\hat{\psi}^+_{k})$ is the standard coordinate transformation matrix from $\{b\}$ to $\{n\}$ and `$-$/$+$' denote the estimation of the corresponding quantity before/after the observation update. $\boldsymbol{w}_{\boldsymbol{a},k}\thicksim\mathcal{N}(0, \boldsymbol{\sigma}^2_{\boldsymbol{a}})$ and $\boldsymbol{w}_{r,k} \thicksim \mathcal{N}(0, \sigma^2_{r})$ are white Gaussian noises. Constant bias components can often be measured and compensated for before missions. Here, we focus on estimating the sensor bias due to bias instability. This component can be modeled as a Gaussian Markov process such that
\begin{equation}
\hat{\boldsymbol{b}}^-_{*,k+1} = (1 - \delta t / \tau_{\boldsymbol{b}}) \, \hat{\boldsymbol{b}}^+_{*,k} + \boldsymbol{w}_{\boldsymbol{b},k}\, \delta t,
\end{equation}
where $(*) \in \{\boldsymbol{a}, r, \boldsymbol{z}\}$ with $\tau_{\boldsymbol{b}}$ being the autocorrelation time constant. The driving process, $\boldsymbol{w}_{b,k} \thicksim \mathcal{N}(0, 2f\boldsymbol{\sigma}^2_{\boldsymbol{b}}/\tau_{\boldsymbol{b}})$, is assumed to be white Gaussian and $f$ is the sampling frequency of the corresponding sensor.

Most experiments and computations of homogeneous, isotropic turbulence have shown  that the velocity distribution at one point is approximately Gaussian~\cite{FungJCH:92a}. Thus, we approximate the unmodeled turbulent flow component as a Gaussian Markov process. Since the ocean current can be considered invariant within a time period on the order of minutes, we correlate changes in turbulence velocity with the vehicle's velocity such that
\begin{equation}\label{eq:dynamicsEnd}
\hat{\boldsymbol{u}}^{\{n\},-}_{c,k+1} = \left[I_{2\times2} - \text{diag}(\hat{\boldsymbol{v}}^{\{n\},+}_{k})\delta t/L_c\right] \, \hat{\boldsymbol{u}}^{\{n\},+}_{c,k} + \boldsymbol{w}_{\boldsymbol{u},k} \, \delta t,
\end{equation}
where $L_c$ is the correlation length-scale of the turbulent flow component and $\boldsymbol{w}_{\boldsymbol{u},k} \thicksim \mathcal{N}(\boldsymbol{0},2k\boldsymbol{\sigma}^2_{\boldsymbol{u}}/L_c)$ is white Gaussian noise with $k$ being the smallest wave number of the turbulence. The values of $L_c$ and $k$ can be obtained based on empirical evidence given the region of navigation.

As the vehicle navigates in the ocean, relative velocities of the ambient flow with respect to the vehicle can be measured. Such measurements are represented in the body-fixed frame as $\boldsymbol{v}_r^b$. In this work, we consider the relative flow velocity measurement to be obtained by an ADCP. It has been widely implemented in many marine engineering applications~\cite{BrumleyBH:1991a} including related current-aided navigation techniques for AUV descending~\cite{Medagoda:16a}. Given the pre-loaded current velocity map $\Phi(\hat{\boldsymbol{p}}^{\{n\}}, t)$, the measurement model can be described as 
\begin{equation}\label{eq:measuremetModel}
\begin{aligned}
\hat{\boldsymbol{z}}_k = h(\hat{\boldsymbol{x}}^-_k, \boldsymbol{w}_{\boldsymbol{z},k}) = R^b_n(\hat{\boldsymbol{\psi}}^-_k)   & \left[ \Phi(\hat{\boldsymbol{p}}^{-}_{k}, t_k) +  \hat{\boldsymbol{u}}^{\{n\},-}_{c,k} -  \right. \\ & \left. \hat{\boldsymbol{v}}^{\{n\},-}_{k} \right] + \hat{\boldsymbol{b}}^-_{\boldsymbol{z},k} + \boldsymbol{w}_{\boldsymbol{z},k},
\end{aligned}
\end{equation}
where $\boldsymbol{w}_{\boldsymbol{z},k} \thicksim \mathcal{N}(\boldsymbol{0}, \boldsymbol{\sigma}^2_{\boldsymbol{z}})$ is the white Gaussian observation noise.

\subsection{Implementation with a Marginalized Particle Filter}\label{sec:implementation}
Based on the aforementioned state vector factorization, the system state vector can be divided into $\boldsymbol{x}^{\text{PF}}_k = \boldsymbol{p}^{\{n\}}_k$ and $\boldsymbol{x}^{\text{KF}}_k = [\boldsymbol{v}^{\{n\}\top}_k, \psi_k, \boldsymbol{b}^{\{b\}\top}_{\boldsymbol{a},k}, b_{r,k},\boldsymbol{b}^{\{b\}\top}_{\boldsymbol{z},k}, \boldsymbol{u}^{\{n\}\top}_{c,k}]^\top$. The linearized system dynamics (\ref{eq:dynamicsStart}--\ref{eq:dynamicsEnd}) and the measurement model (\ref{eq:measuremetModel}) can be compactly expressed as 
\begin{align}
\hat{\boldsymbol{x}}^{\text{PF},-}_{k+1} &= \hat{\boldsymbol{x}}^{\text{PF},+}_{k} + A^{\text{PF}} \, \hat{\boldsymbol{x}}^{\text{KF},+}_{k}, \label{eq:pfstates} \\
\hat{\boldsymbol{x}}^{\text{KF},-}_{k+1} &= F_k(\hat{\boldsymbol{x}}^{\text{KF},+}_{k})\,\hat{\boldsymbol{x}}^{\text{KF},+}_{k} + G_k(\hat{\boldsymbol{x}}^{\text{KF},+}_{k})\,\boldsymbol{w}_{\boldsymbol{x^{\text{KF}}},k}, \label{eq:kfstates} \\
\hat{\boldsymbol{z}}_{k} &= H_k(\hat{\boldsymbol{x}}^{\text{PF},-}_{k}) \, \hat{\boldsymbol{x}}^{\text{KF},-}_{k} + \boldsymbol{w}_{\boldsymbol{z},k},
\end{align}
where $A^{\text{PF}} = \text{diag}(\delta t I_{2\times 2}, 0_{8 \times 8})$,
\begin{equation*}
F_k = \begin{bmatrix}
I_{2 \times 2} & F_{12} & F_{13} & 0_{2\times 1} & 0_{2\times 2} & 0_{2\times 2} \\
0_{1\times 2} & 1 & 0_{1\times 2} & -\delta t & 0_{1\times 2} & 0_{2\times 2} \\
0_{2\times 2} & 0_{2\times 1} & F_{33} & 0_{2\times 1} & 0_{2\times 2} & 0_{2\times 2}\\
0_{1\times 2} & 0 & 0_{1\times 2} & F_{44} & 0_{1\times 2} & 0_{1\times 2}\\
0_{2\times 2} & 0_{2\times 1} & 0_{2\times 2} & 0_{2\times 1} & F_{55} & 0_{2\times 2}\\
F_{61} & 0_{2\times 1} & 0_{2\times 2} & 0_{2\times 1} & 0_{2\times 2} & F_{66}
\end{bmatrix},
\end{equation*}
with $F_{12} = \nabla_{\psi}R^n_b(\hat{\psi}^+_k) (\boldsymbol{a}_k + \hat{\boldsymbol{b}}^{+}_{\boldsymbol{a},k})\delta t$, $F_{13} = -R^n_b(\hat{\psi}^+_k)\delta t$, $F_{33} = (1-{\delta t}/{\tau_{\boldsymbol{a}}})I_{2\times 2}$, $F_{44} = 1-{\delta t}/{\tau_{r}}$, $F_{55} = (1-{\delta t}/{\tau_{\boldsymbol{z}}})I_{2\times 2}$, $F_{61} = {\text{diag}\{-\hat{\boldsymbol{u}}^+_{c,k}\}}/{L_c}\delta t$, $F_{66} = I_{2\times 2} - \text{diag}\{\hat{\boldsymbol{v}}^{+}_k\}\delta t/L_c$, $G_k = \text{diag}\{-R^n_b(\hat{\psi^+_k})\delta t, -\delta t, \delta tI_{7\times 7}\}$, and $H_k = \text{diag}\{-R^b_n(\hat{\psi}^-_k), \nabla_\psi R^b_n(\hat{\psi}^-_k)[\Phi(\hat{\boldsymbol{p}}^-_k, t_k) + \hat{\boldsymbol{u}}^-_{c,k} - \hat{\boldsymbol{v}}^-_k], 0_{2\times 3}, I_{2\times 2}, R^b_n(\hat{\psi}^-_k)\}$.

For tracking problems in vehicle navigation, prior distributions of the system's states are often known. Therefore, updated system states can be sequentially estimated as new sensor measurements are acquired. At each time step, $N_p$ samples are drawn from the proposal distribution (prior distribution in this case) based on (\ref{eq:pfstates}) as an estimate of $\boldsymbol{x}^{\text{PF},-}_{k+1}$, during which  $\boldsymbol{x}^{\text{KF},+}_{k}$ are considered as process noise. Each sample maintains an individual estimate of the remaining states $\boldsymbol{x}^{\text{KF},i,-}_{k+1}$ with $i = 1\cdots N$. The estimates of the remaining states propagate as Gaussian distributions according to (\ref{eq:kfstates}), the covariance matrix of which are updated accordingly as
\begin{equation}\label{eq:P_update}
P^{i,-}_{k+1} = F^i_kP^{i,+}_{k}(F^i_k)^\top + G^i_k Q (G^i_k)^\top.
\end{equation}
Matrix $Q = E\{\boldsymbol{w}_{\boldsymbol{x^{\text{KF}}}}, \boldsymbol{w}^\top_{\boldsymbol{x^{\text{KF}}}}\}$ is the covariance of the process noise for $\boldsymbol{x}^{\text{KF}}_{k}$. 

When a new measurement of the relative flow velocity, $\boldsymbol{z}_{k+1}$, is obtained, which typically occurs less frequently than the system state propagation process, the sample weight, $w^i_{k}$, associated with each particle, $\boldsymbol{x}^{\text{PF},i,-}_{k}$, updates based on a Gaussian likelihood as previously discussed with (\ref{eq:weights})
\begin{equation}\label{eq:updateW}
\begin{aligned}
w^i_{k} \propto w^i_{k-1} \dfrac{1}{\sqrt{2\pi\det(S^i_{k})}} &\exp \left\{ -{1}/{2}  (\boldsymbol{z}_{k} - h(\hat{\boldsymbol{x}}^{i,-}_k,0))^\top \right. \\ & \left. (S^i_{k})^{-1}(\boldsymbol{z}_{k} - h(\hat{\boldsymbol{x}}^{i,-}_k,0))\right\}.
\end{aligned}
\end{equation}
Matrix $S^i_{k}$ represents the covariance of the innovation $\boldsymbol{e}^i_k = \boldsymbol{z}_{k} - h(\hat{\boldsymbol{x}}^{i,-}_k,0)$ and it is calculated based on the relationship
\begin{equation}\label{eq:innovationCov}
S^i_k = H^i_k P^{i,-}_k (H^i_k)^\top + R,
\end{equation}
where $R = E\{\boldsymbol{w}_{\boldsymbol{z}}, \boldsymbol{w}^\top_{\boldsymbol{z}}\}$ is the covariance of the observation uncertainty. Estimates of the remaining states are then corrected using the new observation based on
\begin{align}
K^i_k &= P^{i,-}_k(H^i_k)^\top(S^i_k)^{-1}, \\
\hat{\boldsymbol{x}}^{\text{KF},i,+}_k &=  \hat{\boldsymbol{x}}^{\text{KF},i,-}_{k} + K^i_k\boldsymbol{e}^i_k, \\
P^i_k &= (I - K^i_kH^i_k)P^{i,-}_k. \label{eq:Cov}
\end{align}
Algorithm \ref{alg:1} summarizes the current-aided inertial navigation framework.
\begin{algorithm}
\caption{Current-aided Inertial Navigation System}
\label{alg:1}
\begin{algorithmic}[1]
\Require{$\boldsymbol{x}_0$, $\boldsymbol{\Phi}(\boldsymbol{p},t)$, $\boldsymbol{a}_{0:k}$, and $r_{0:k}$}

	\State Sample $N_p$ particles $\{\boldsymbol{x}^{i}_0, w^i_0\} \sim \Pr(\boldsymbol{x}^{\text{PF}}_0)$ and set $w^i_0 = 1/N_p \quad \forall \quad i = 1,\cdots N_p$ 
	\State Initialize $\hat{\boldsymbol{x}}^{\text{KF},i}_0$ and $P^i_0$
\While{$t_k < t_f$}
	\For{$i = 1,2,\cdots,N_p$}
		\State Predict the relative flow velocity $\hat{\boldsymbol{z}}^i_k$ \Comment{\eqref{eq:measuremetModel}}
		\State Update the particle weight $w^i_k$ based on the actual sensor measurement $\boldsymbol{z}_k$ \Comment{\eqref{eq:updateW}}
		\State Update $\hat{\boldsymbol{x}}^{\text{KF},i}_k$ and $P^i_k$ \Comment{\eqref{eq:innovationCov}--\eqref{eq:Cov}}
	\EndFor
	\State Normalize particle weights such that $\sum^{N_p}_{i=1}{w^i_k} = 1$
	\If{$N_\text{eff} < N_p/2$} \Comment{$N_\text{eff} = \left( \sum^N_{i=1} (w^i_k)^2 \right)^{-1}$}
		\State Resample particles with replacements $N_p$ times
	\EndIf
	\For{$i = 1,2,\cdots,N_p$}
		\State Predict $\hat{\boldsymbol{x}}^{i}_{k+1}$ and $P_{k+1}$ \Comment{\eqref{eq:pfstates}--\eqref{eq:kfstates}}
	\EndFor
\EndWhile
\end{algorithmic}
\end{algorithm}

\subsection{Parametric Cram\'{e}r-Rao Lower Bound (CRLB)}\label{sec:CRLB}
Before testing the practical performance of the current-aided inertial navigation system, it is beneficial to estimate the expectable theoretical performance of an unbiased estimator for the system of interest. To this end, we analyze the theoretical estimation bound, using the parametric CRLB, of the reduced system dynamics. The reduction is in regard to omitting the sensor bias and the turbulent flow component from the original system state vector $\boldsymbol{x}_k$. This should not prevent the validity of the resulting performance bound in evaluating the performance of the proposal method since a tighter bound can be expected when all states are included. When the accuracy of trajectory-following and the performance of disturbance rejection can be guaranteed, this analysis can also be used as an indication of the possible navigation performance to be expected given a current map and a prescribed vehicle trajectory.

We consider the reduced system dynamics
\begin{align}
\widetilde{\boldsymbol{x}}^-_{k+1} &= \widetilde{F}^0_k \widetilde{\boldsymbol{x}}^-_{k} + \widetilde{G}^0_k\widetilde{\boldsymbol{w}}_{\boldsymbol{x},k},\\
\boldsymbol{z}_k &= \widetilde{H}^0_k\widetilde{\boldsymbol{x}}^-_k + \boldsymbol{w}_{\boldsymbol{z},k},
\end{align}
where $\widetilde{\boldsymbol{x}}_{k} = [\boldsymbol{p}^{\{n\}\top}_k, \boldsymbol{v}^{\{n\}\top}_k, \psi_k]^\top$ and $\widetilde{\boldsymbol{w}}_{\boldsymbol{x},k} = [\boldsymbol{w}_{\boldsymbol{a}}^\top, w_r]^\top \sim \mathcal{N}(\boldsymbol{0}, \widetilde{Q})$. Superscript `0' indicates that the corresponding variable is evaluated with the true states and their actual rates of changes. The Jacobian matrices take the forms
\begin{align*}
&\widetilde{F}^0_k = \begin{bmatrix}
I_{2\times 2} & \delta t I_{2\times 2} & 0_{2\times 1} \\
0_{2\times 2} & I_{2\times 2} & \nabla_{\psi}R^n_b \boldsymbol{a}^0_k \delta t \\
0_{1\times 2} & 0_{1\times 2} & 1
\end{bmatrix}, \\
&\widetilde{G}^0_k = \begin{bmatrix}
0_{2\times 2} & 0_{2\times 1}\\
-R^n_b\delta t & 0_{2\times 1}\\
0_{1\times 2} & -\delta t
\end{bmatrix},
\widetilde{H}^0_k = \begin{bmatrix}
R^b_n \nabla_{\boldsymbol{p}}\Phi(\boldsymbol{p}^0_k,t_k) \\ -R^b_n \\ \nabla_\psi R^b_n[\Phi(\boldsymbol{p}^0_k,t_k) - \boldsymbol{v}^0_k]
\end{bmatrix}^\top.
\end{align*}
The parametric CRLB for one-step-ahead prediction and filtering~\cite{BergmanN:1999a} can then be computed recursively based on
\begin{align}
\widetilde{P}_{k+1|k} &= \widetilde{F}^0_k\widetilde{P}_{k|k} \widetilde{F}^{0,\top}_k + \widetilde{G}^0_k\widetilde{Q}\widetilde{G}^{0,\top}_k,\\
\widetilde{P}_{k|k} &= \widetilde{P}_{k|k-1} - \\ &\widetilde{P}_{k|k-1}\widetilde{H}^{0,\top}_k(\widetilde{H}^0_k\widetilde{P}_{k|k-1}\widetilde{H}^{0,\top}_k + R)^{-1}\widetilde{H}^0_k\widetilde{P}_{k|k-1}. \label{eq:CRLBfilter}
\end{align}
It is worth mentioning that the CRLB is a function of the spatial gradient of the mean flow component. Relationship (\ref{eq:CRLBfilter}) indicates a smaller positioning variance when the current map has larger spatial variations.

\section{Performance Evaluation in Turbulent Flows}\label{sec:simulation}
We demonstrate the performance of the current-aided navigation scheme in simulations with two different background flow fields: a double-gyre flow field~\cite{Shadden:05a} and a meandering jet flow field~\cite{Samelson:92a}. They both resemble typical, real-world ocean flow patterns and have been widely used to study the transportation and mixing properties of ocean flows due to their simple analytical expressions. Additional turbulent components that resemble the energy spectrum of real-world turbulence were superposed on top of the mean flow. The lawn-mowing vehicle trajectory, a common trajectory widely adopted in ocean sampling tasks~\cite{SmithR:11a}, was used in both cases. State estimation for vehicles following such a trajectory is quite challenging since there exist multiple large-angle turns, which may lead to large attitude estimation error and further affect the overall navigation performance. The lawn-mowing trajectory was locally perturbed for simulating a more realistic scenario where the vehicle has a bounded trajectory-following error.

\subsection{Sensor Sample Generation}\label{sec:INSmodel}
Given a prescribed vehicle trajectory, actual vehicle acceleration $\boldsymbol{a}^{\{n\}}_{\text{real}}$ and angular velocity $r_{\text{real}}$ were calculated. In order to generate noisy measurement samples from an INS, additional noise terms were injected. Depending on the manufacture and the class, an INS typically has different sources of errors due to temperature fluctuation, calibration errors, constant bias, random walk, bias instability, etc. Here, we assume that the INS in use has a built-in temperature sensor to compensate for the measurement noise caused by temperature fluctuation, and the constant bias can also be estimated and compensated for before deployment. 

We corrupted the vehicle's actual acceleration in the body frame and angular velocity with noises due to random walk and bias instability such that
\begin{align}
\boldsymbol{a}^{\{b\}}_{\text{noisy}} &= R^b_n(\psi_{\text{real}})\boldsymbol{a}^{\{n\}}_{\text{real}} + \boldsymbol{b}^{\boldsymbol{a}}_{\text{instability}} + \boldsymbol{\nu}^{\boldsymbol{a}}_{\text{white}}\,,\\
r_{\text{noisy}} &= r_{\text{real}} + \boldsymbol{b}^r_{\text{instability}} + \boldsymbol{\nu}^r_{\text{white}}\,,
\end{align}
where $\boldsymbol{\nu}^*_{\text{white}}$ is white Gaussian noise, and $\dot{\boldsymbol{b}}^*_{\text{instability}} = -{\boldsymbol{b}}^*_{\text{instability}}/\tau_* + \boldsymbol{w}_{\boldsymbol{b}}$ is an exponentially correlated process driven by a white Gaussian noise $\boldsymbol{w}_{\boldsymbol{b}}$ with standard deviation (SD) $\boldsymbol{\sigma}_{\boldsymbol{b}}$ and correlation time $\tau_*$. The standard deviation of the white Gaussian driving process for bias instability can be determined based on the resultant power spectrum density $Q_{\boldsymbol{\nu}} = 2\boldsymbol{\sigma}_{\boldsymbol{b}}^2/\tau_*$. Sensor noise generated in such a way leads to Allen variance similar to corresponding physical sensors. 

Similarly, we corrupted the real, relative background flow velocity (including both the mean flow and the turbulent components) with respect to the vehicle with random walk and bias instability modeled as a first-order Markov process driven by white Gaussian noise. The resulting noisy relative flow velocities are used as actual observation measurements in the following analysis.
 
In the following two tests, parameters for sensor sample generation were based on characteristics of an automotive-grade INS \textit{VN-100} from VectorNav and an \textit{RDI 1200 Khz} DVL with ADCP function from Teleldyne. Values of the related parameters used in this work are tabulated in Table~\ref{tb:INSparameter}.
\begin{table}
\centering
\caption{Parameters for sensor sample generation used in simulations with artificial turbulent flow fields.}
\begin{tabular}{l  l}
\toprule[0.75mm]
INS  & VectorNav \textit{VN-100}\\
\hline
Accel. bandwidth & 260 Hz \\
Accel. white noise SD ($\sigma_a$) & $0.14$ mg/$\sqrt{\text{Hz}}$ \\
Accel. in-run bias stability ($\sigma_{b_a}$) & 0.04 mg \\
Accel. correlation time constant ($\tau_a$) & 300 s \\
Gyro. bandwidth & 256 Hz \\
Gyro. white noise SD ($\sigma_r$) & 0.0035 ${}^\circ$/s/$\sqrt{\text{Hz}}$ \\
Gyro. in-run bias stability ($\sigma_{b_r}$) & 10 ${}^\circ$/hr \\
Gyro. correlation time constant ($\tau_r$) & 300 s \\
Update rate used & 10 Hz \\
\midrule[0.5mm]
ADCP  & RDI \textit{1200 kHz}\\
\hline
Measurement uncertainty ($\sigma_{\text{z}}$) & 0.01 m/s\\
Bias instability ($\sigma_{b_z}$) & 0.01 m/s \\
Correlation time constant ($\tau_z$) & 100 s \\
Update rate used & 1 Hz \\
\bottomrule[0.75mm]
\end{tabular}
\label{tb:INSparameter}
\end{table}

\subsection{Turbulence with Kinematic Simulation (KS)}\label{sec:KS}
The current-aided inertial navigation system depends on the knowledge of the background flow field. The fact that small-scale turbulence cannot be fully resolved by general circulation models poses a potential issue to the navigation performance of a vehicle. In order to evaluate the robustness of the proposed technique in turbulent flow fields, it is necessary to create background flows with characteristics that resemble real-world ocean flows. To this end, additional turbulent components are superposed on top of the mean flow that is assumed to be known by the vehicle.

To create a turbulent flow field with consistent length-scales between the mean flow and local perturbations, small-scale turbulence was generated through Kinematic Simulation (KS)~\cite{FungJCH:92a}. The KS models are non-Markovian Lagrangian models for turbulent-like flow structures widely adopted in investigation of particle dispersion/collision where kinetic interactions do not play a key role in the analysis. Although the resulting flow fields do not necessarily satisfy the dynamic equations, flow field generated with KS models have shown good agreement with experimental measurements in terms of Lagrangian statistics. Despite their simple mathematical forms, the solutions have a self-similar energy spectrum over a large range of scales, making it a good tool for introducing small-scale eddies to a mean flow (large-scale eddies).

\begin{figure}
\centering
\includegraphics[width = 0.27\linewidth]{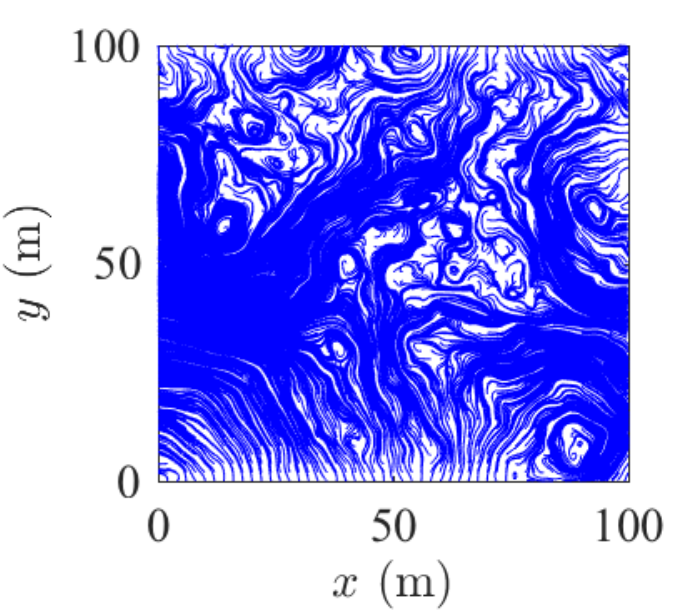}
\includegraphics[width = 0.355\linewidth]{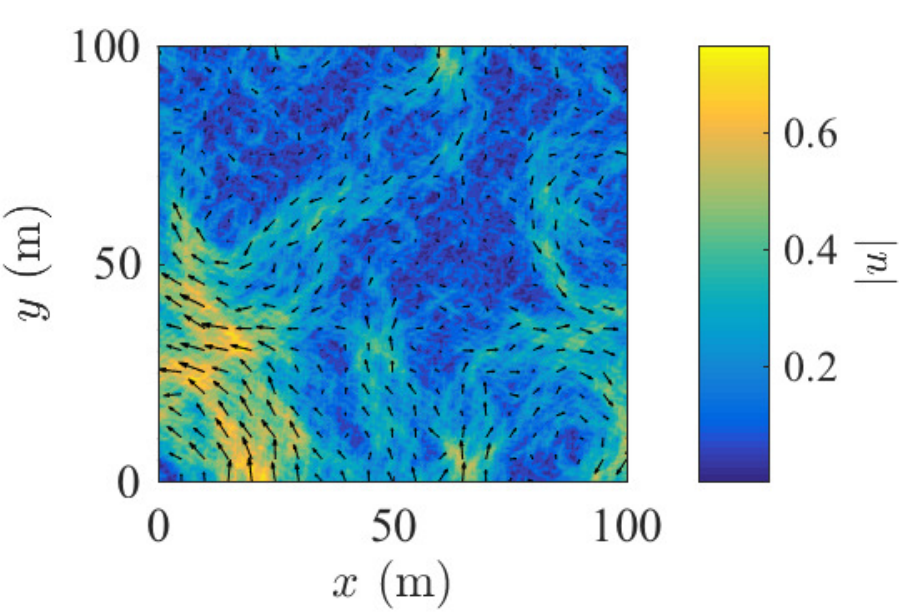}
\caption{Turbulent flow field generated with KS: streamlines (left); velocity field (right).  Arrows indicate the direction of the flow and their length is proportional to the flow speed.}
\label{fig:KS}
\end{figure}

The temporal structure of the turbulent flow component is determined by a frequency series generated using the kinetic simulation inertial model (KSIM)~~\cite{FungJCH:92a}. The flow field can be calculated as a summation across a series of modes
\begin{equation}
\begin{aligned}
\boldsymbol{u}(\boldsymbol{x},t) = \sum^{N_k}_{n=1}\left[  (\boldsymbol{a}_n  \times  \hat{\boldsymbol{k}}_n) \cos (\boldsymbol{k}_n \cdot \boldsymbol{x} + \omega_n t)  \right.\\
\left. + (\boldsymbol{b}_n\times \hat{\boldsymbol{k}}_n)\sin (\boldsymbol{k}_n\cdot \boldsymbol{x} + \omega_n t) \right],
\end{aligned}
\end{equation}
where $N_k$ is the number of modes in turbulent simulation dictating the complexity of the resulting flow field. It is constrained to have a Kolmogorov-like energy spectrum 
\begin{equation}
E(k) = \left\{ \begin{array}{rcl}
\alpha_k\varepsilon^{2/3}k^{-5/3} & \text{for} & k_c\leq k \leq k_\eta\\
0 & \text{for} & \text{otherwise}
\end{array}\right.,
\end{equation}
where $k$ represents the wavenumber, $\alpha_k = 1.5$ is the Kolmogorov constant~\cite{GrantHL:1962,GibsonMM:1963a}, and $\varepsilon$ is the rate of dissipation of kinetic energy per unit mass. Two boundary values for the wavenumber are the cut-off wavenumber, $k_c$, and the maximum simulated wavenumber, $k_\eta$. Given a flow velocity variance of the large-scale component, $\langle\boldsymbol{u}^2_{ls}\rangle$, the rate of kinetic energy dissipation can be calculated as
\begin{equation}
\varepsilon = \left[ \dfrac{\langle\boldsymbol{u}^2_{ls}\rangle}{\alpha_k}\left( \dfrac{1}{k^{2/3}_c} - \dfrac{1}{k^{2/3}_\eta} \right)^{-1} \right]^{3/2}.
\end{equation}
We generated the wavenumber series based on a geometric distribution such that $k_n = k_c(L/\eta)^{(n-1)/(N_k-1)}$, where $L = 2\pi/k_c$ was chosen as the correlation length-scale and $\eta = 2\pi/k_\eta$. Therefore, the frequency series can be computed based on the desired energy spectrum as $\omega_n = \sqrt{({k_n^3}/{\alpha_k})E(k_n)}$.
Finally, random vectors $\boldsymbol{a}_n$, $\boldsymbol{b}_n$ and $\boldsymbol{k}_n$ are determined based on mutually uncorrelated random angles $\phi_n$ such that $\boldsymbol{a}_n = a_n (\cos\phi_n, -\sin \phi_n)$, $\boldsymbol{b}_n = b_n (-\cos \phi_n, \sin \phi_n)$, $\boldsymbol{k}_n = k_n (\sin \phi_n, \cos \phi_n)$, where $a_n^2 = b_n^2 = E(k_n)\Delta k_n$, and $\hat{\boldsymbol{k}}_n = \boldsymbol{k_n}/k_n$. Fig.~\ref{fig:KS} shows an example flow region generated using this method with parameters $L = 200$ m, $\eta = 0.001$ m, and $N_k = 100$.

\subsection{Double-Gyre Flow Field}
\begin{figure}
\centering
\includegraphics[width = 0.75\linewidth]{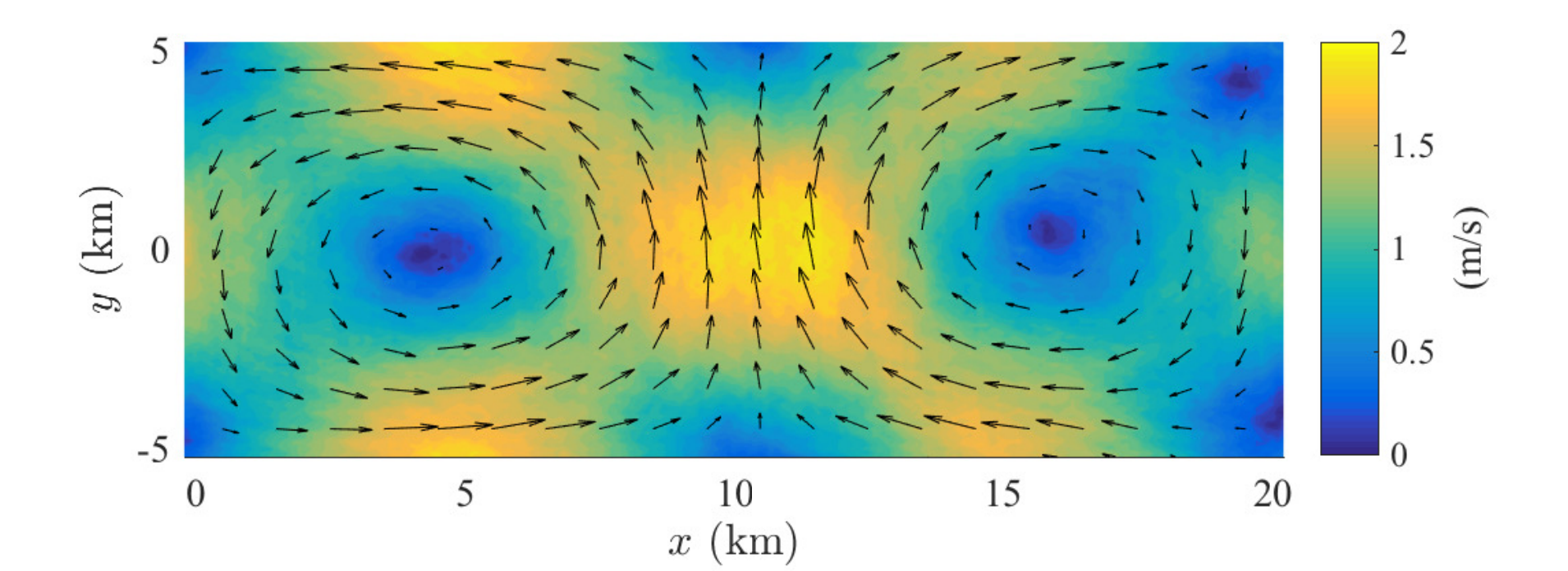}
\caption{Double-gyre flow field with KS turbulence at $t = 0$.  Arrows indicate the direction of the flow and their length is proportional to the flow speed.}
\label{fig:DGflow}
\end{figure}

In the first test case, we considered the navigation problem in a time-dependent double-gyre flow field. The double-gyre flow represents a typical large-scale ocean circulation phenomenon often observed in the northern mid-latitude ocean basins. It is quite dominant and persistent in oceans and consists of a sub-polar and a sub-tropical gyres. The flow velocity of a double-gyre flow field is defined by the stream function
\begin{equation}
\phi(x,y,t) = A*\sin(\pi f(x,t)) \sin(\pi y),
\end{equation}
where the time dependency is introduced by
\begin{align}
f(x,t) &= a(t)x^2+b(t)x,\nonumber \\
a(t) = \epsilon \sin (\omega t)& \quad \text{and} \quad
b(t) = 1-2\epsilon \sin(\omega t), \nonumber
\end{align}
over a nondimensionalized domain of $[0,2] \times [0,1]$. Here, $\epsilon$ dictates the magnitude of oscillation in the $x$-direction, $\omega$ is the oscillation period, and $A$ controls the velocity magnitude. The resulting velocity field can be calculated based on $[u = -\partial \phi/\partial y, v = \partial \phi/\partial x]$. We chose $A = 1.5/\pi$, $\epsilon = 0.3$, $\omega = 2\pi$ and applied a length-scale of $L = 10$ km to create a flow field filling the domain $[0,20] \times [-5,5]$ km with a maximum current speed of 1.5 m/s. The resulting flow field superposed with random turbulent components at $t = 0$ is shown in Fig.~\ref{fig:DGflow}.

\begin{figure}
\centering
\includegraphics[width = 0.7\linewidth]{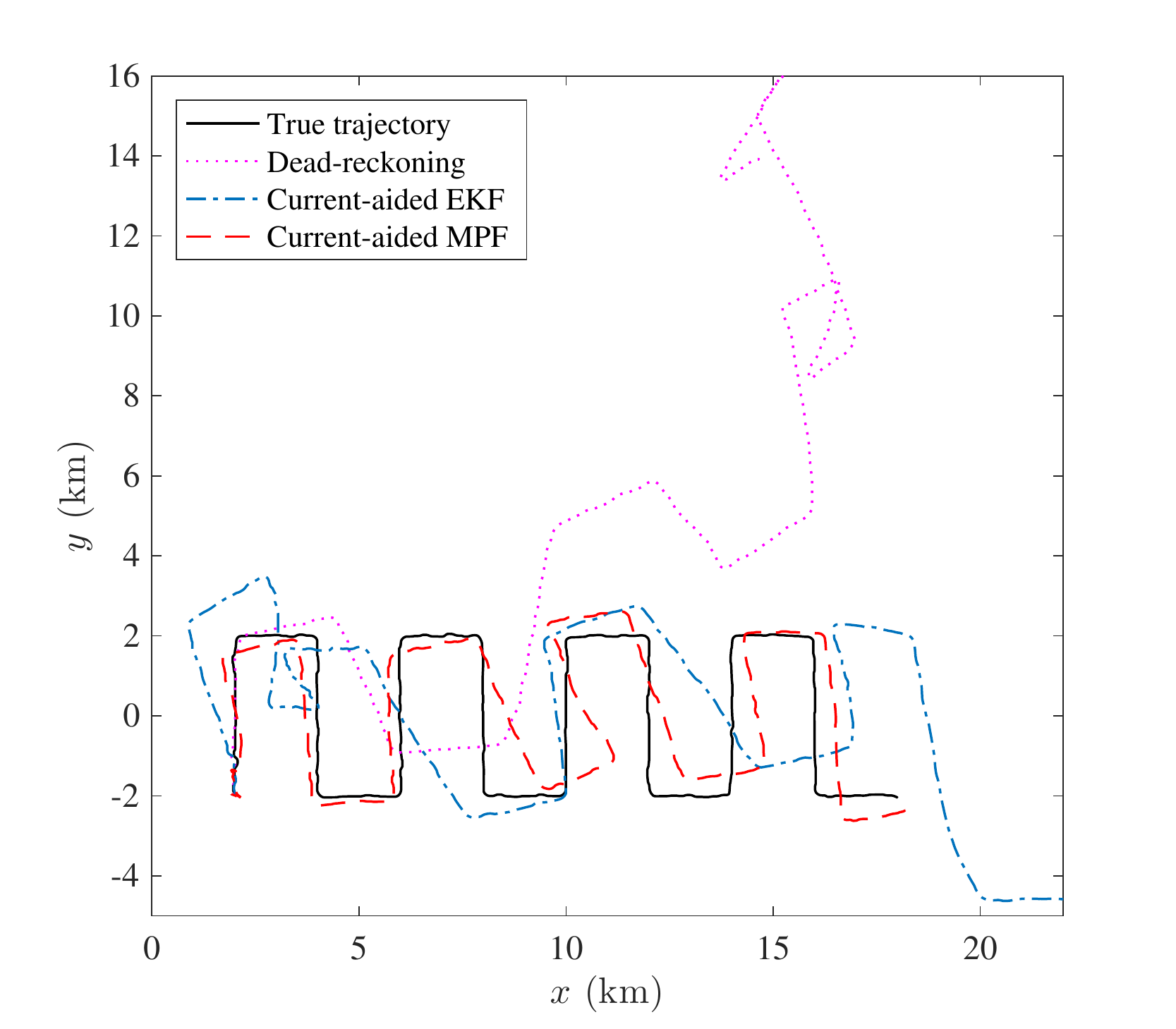}
\caption{Comparison between real and estimated vehicle paths in a turbulent double-gyre flow field.}
\label{fig:DGpath}
\end{figure}

Fig.~\ref{fig:DGpath} compares the navigation results under different estimation schemes against the actual vehicle path from one simulation run in the turbulent double-gyre flow field. The dead-reckoning scheme simply integrates noisy INS samples generated in Sec.~\ref{sec:INSmodel}. Such an estimate severely deviates from the true path as expected. The current-aided EKF scheme is based on an accurately known initial vehicle state. The EKF was only applied to $\boldsymbol{x}^{\text{KF}}$ and the vehicle's position propagates deterministically using the estimated vehicle velocity. Such an estimator can be considered as a degenerated MPF estimator with a single particle. Although knowledge of the general ocean circulation helps reduce the dead-reckoning error, it can be observed that this fully-parametric estimator will still suffer from large, diverging drifting error. On the other hand, the proposed current-aided MPF results in more consistent estimation, where a total number of 100 particles were initialized with EKF states:
\begin{align*}
&\boldsymbol{v}^{[i]}_0 \sim \mathcal{N}(\boldsymbol{v}^\text{true}_0,\, 10^{-6}I_{2\times1})\,\text{m/s}, \quad \psi^{[i]}_0 \sim \mathcal{N}(\psi^\text{true}_0,\, 10^{-8}),\\
&\boldsymbol{b}^{[i]}_{\boldsymbol{a},0} = \mathbf{0}_{2\times1}, \quad b^{[i]}_{r,0} = 0, \quad \boldsymbol{b}^{[i]}_{\boldsymbol{z},0} = \mathbf{0}_{2\times1}, \quad \boldsymbol{u}^{[i]}_{c,0} =  \mathbf{0}_{2\times1},
\end{align*}
and at random locations of $\boldsymbol{p}^{[i]}_0 \sim \mathcal{N}(\boldsymbol{p}^\text{true}_0,\, 10^6I_{2\times1})$~m.

\begin{figure}
\centering
\includegraphics[width = 0.7\linewidth]{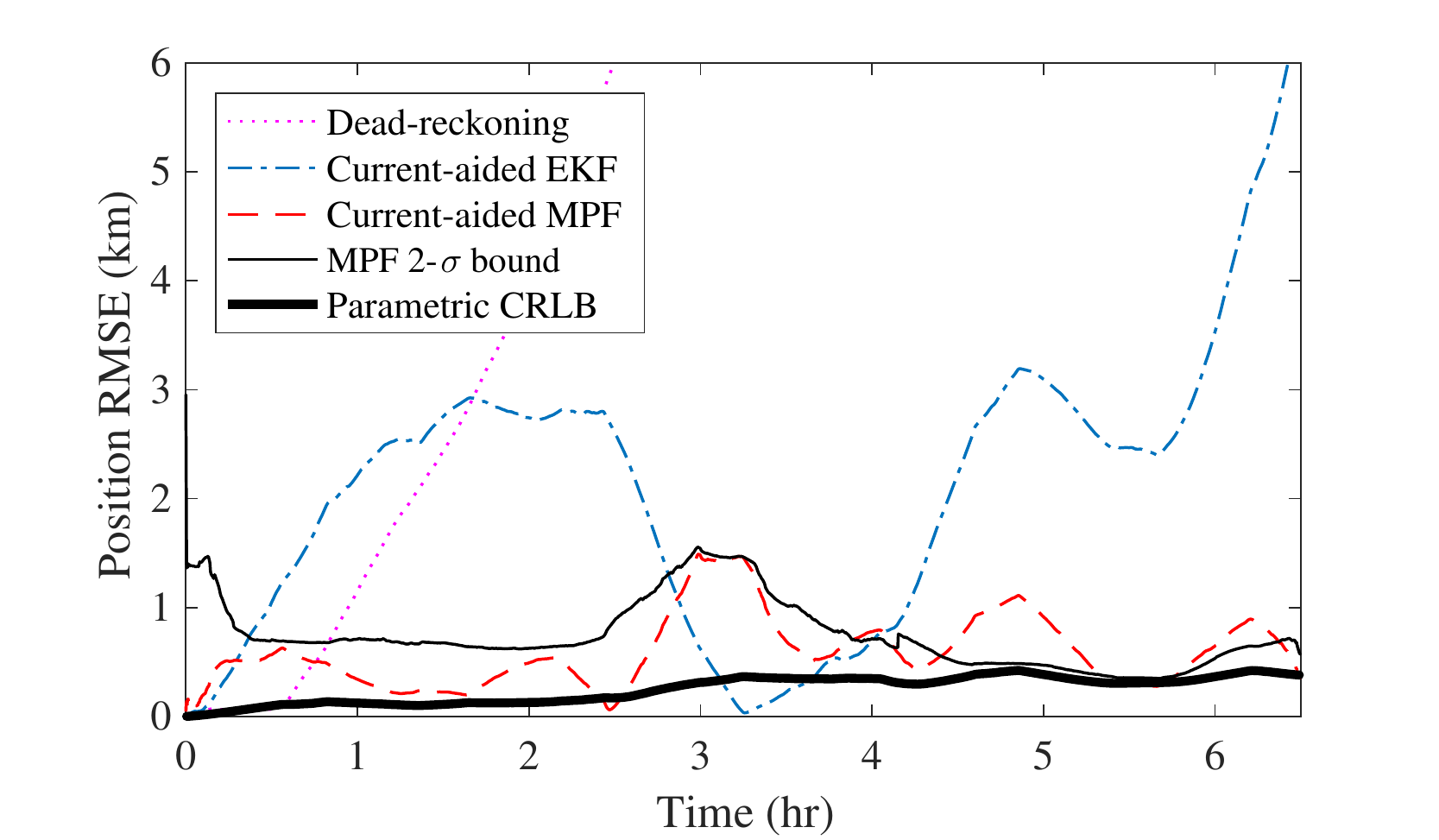}
\caption{Position RMSE and parametric CRLB for a vehicle navigating in a turbulent double-gyre flow field. Results were averaged based on 50 Monte Carlo simulations with random particle initialization and random KS turbulent flow components.}
\label{fig:DGpRMSE}
\end{figure}

\begin{figure}
\centering
\includegraphics[width = 0.7\linewidth]{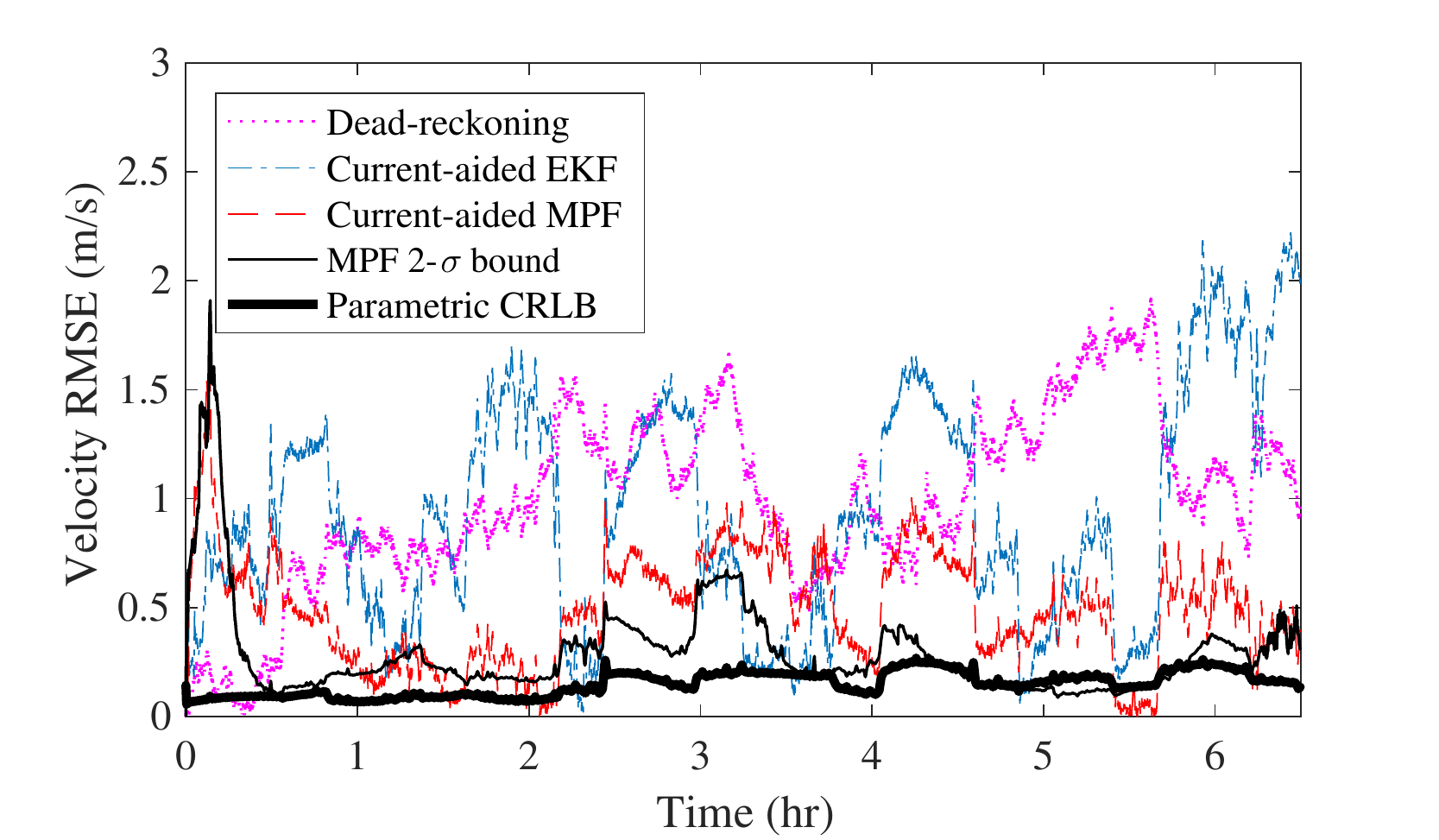}
\caption{Vehicle velocity RMSE and parametric CRLB for a vehicle navigating in turbulent double-gyre flow field.}
\label{fig:DGvRMSE}
\end{figure}

The performance of current-aided MPF can be further evaluated in detail from Fig.~\ref{fig:DGpRMSE} and Fig.~\ref{fig:DGvRMSE}. The root mean square error (RMSE) of vehicle position and velocity estimates were evaluated based on 50 Monte Carlo simulations with random KS turbulent components. The $2$--$\sigma$ estimation bound was compared against the parametric CRLB described in Sec.~\ref{sec:CRLB}. It can be observed that the proposed current-aided MPF converges towards CRLB asymptotically for both position and velocity estimates, indicating that near optimal estimation performance is achieved. It should also be mentioned that the current-aided MPF seems to become over confident as the RMSE occasionally exceeds the $2$--$\sigma$ bound. This is mostly due to particle approximation error of the confidence region. This issue can be remedied by increasing the particle filter's sample size when computational resources permit.

\subsection{Meandering Jet Flow Field}
To further evaluate the performance of the current-aided navigation method, the MPF estimator was tested in a flow field based on a meandering jet model. We consider it to be more of a challenging flow field for the proposed current-aided inertial navigation system. In contrast to the double-gyre flow field, strong currents are concentrated at a small fraction of the domain in a meandering jet model. A large portion of the flow field has zero flow velocity, resulting in a much lower signal-to-noise ratio (SNR) when unmodeled turbulent components are considered. In addition, it contains more repetitive flow features compared to the previous case, which can easily lead to multiple high-likelihood hypotheses.

A striking pattern of vertical and cross-stream flow motions were first observed in the Gulf Stream by Bower and Rossby, who claimed that these motions were closely related to the meandering jet \cite{BowerAS:89a,BowerAS:91a}. Since then, a large body of literature has emerged inspired by their work \cite{SamelsonRM:06a}. Caruso et al.~introduced a meandering current mobility model with a series of investigations on drifters' Lagrangian mobility under the impact of the meandering jet \cite{CarusoA:08a}. The recent prevalence of the meandering jet in underwater sensor network and oceanography studies is mainly resulted from the fact that it includes both major ocean circulation patterns, i.e. currents and large-scale vortices.

\begin{figure}
\centering
\includegraphics[width = 0.75\linewidth]{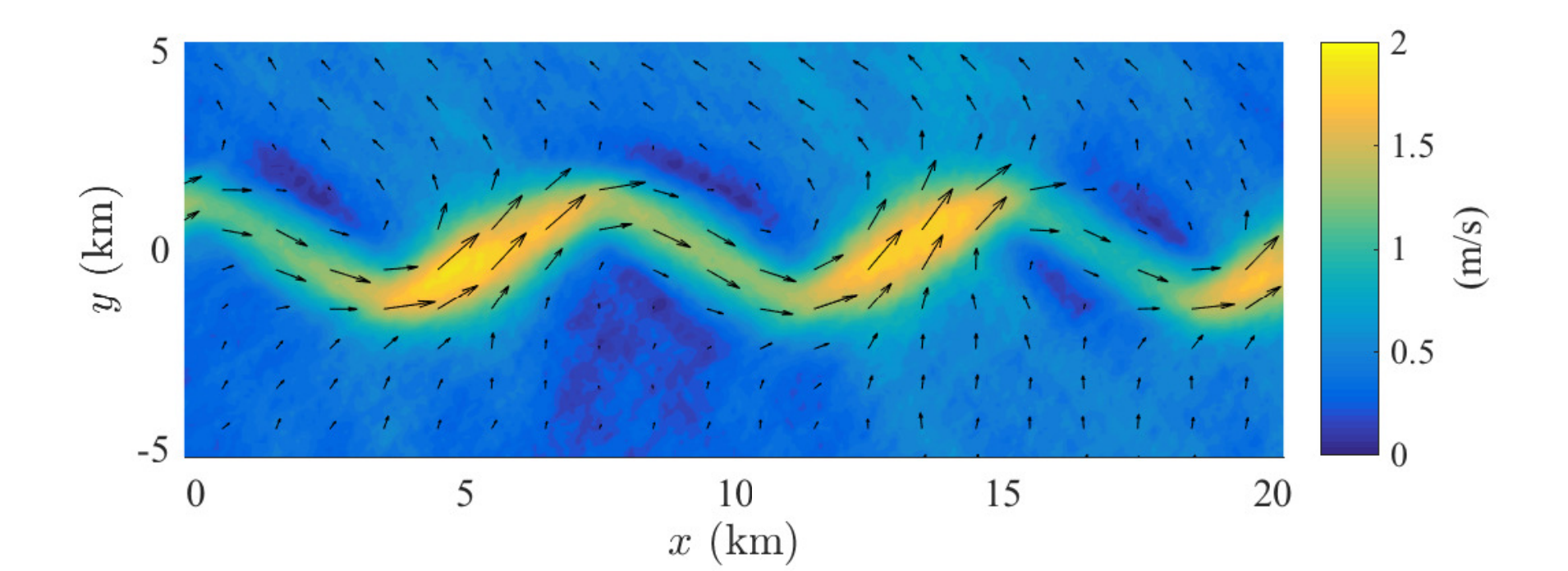}
\caption{Meandering jet flow field with KS turbulence.  Arrows indicate the direction of the flow and their length is proportional to the flow speed.}
\label{fig:MJflow}
\end{figure}

The non-dimensional stream-function of the meandering jet model~\cite{BowerAS:89a} can be expressed as
\begin{align}\label{eq:mj}
\phi(x,y,t) =1 -\tanh\left[ \dfrac{y-B(t)\sin(k(x-c\,t))}{\sqrt{1+k^2B^2(t)\cos^2(k(x-c\,t))}} \right],
\end{align}
where $ B(t) = A+\varepsilon \cos(\omega t)$, $c$ denotes the downstream current phase speed, $k$ determines the number of meanders in a unit length, $A$ is the average meander length-scale, $\varepsilon$ is the magnitude of the meanders, and $\omega$ is the meander frequency. We chose the non-dimensional parameters as follows
\begin{equation*}
A = 1.2, \,\, c = 0.12, \,\, k = 2\pi/7.5, \,\, \omega = 0.4, \,\, \varepsilon = 0.3.
\end{equation*}
With the length-scale being $L = 1$ km and the time-scale being $T = 0.03$ day, the meanders are 7.5 km in size and the maximum flow speed in the meandering jet is approximately 1.5 m/s. The resulting velocity field with the aforementioned parameters superposed with KS turbulence is shown in Fig.~\ref{fig:MJflow}. 

\begin{figure}
\centering
\includegraphics[width = 0.7\linewidth]{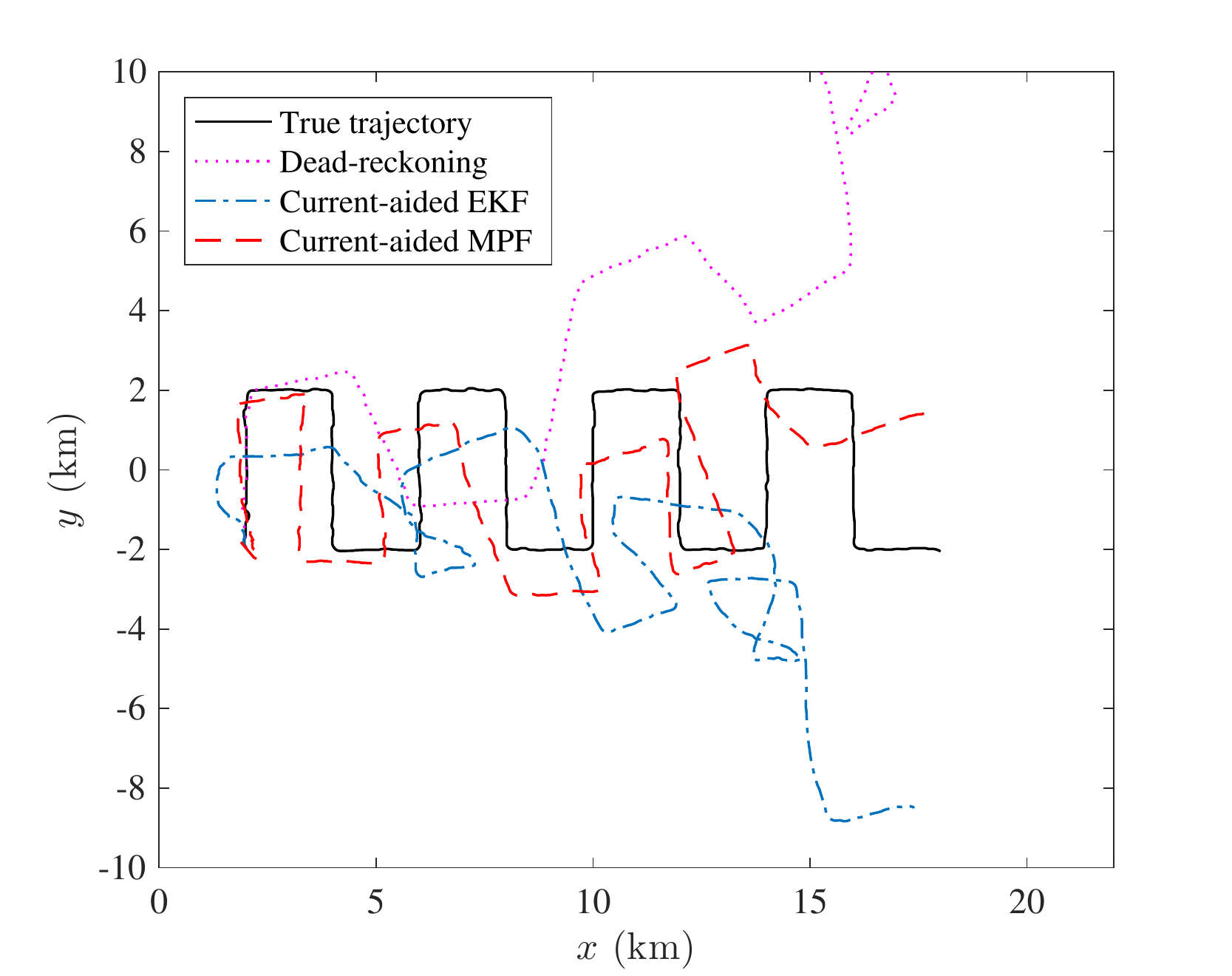}
\caption{Comparison between real and estimated vehicle paths in a turbulent meandering jet flow field.}
\label{fig:MJpath}
\end{figure}

\begin{figure}
\centering
\includegraphics[width = 0.7\linewidth]{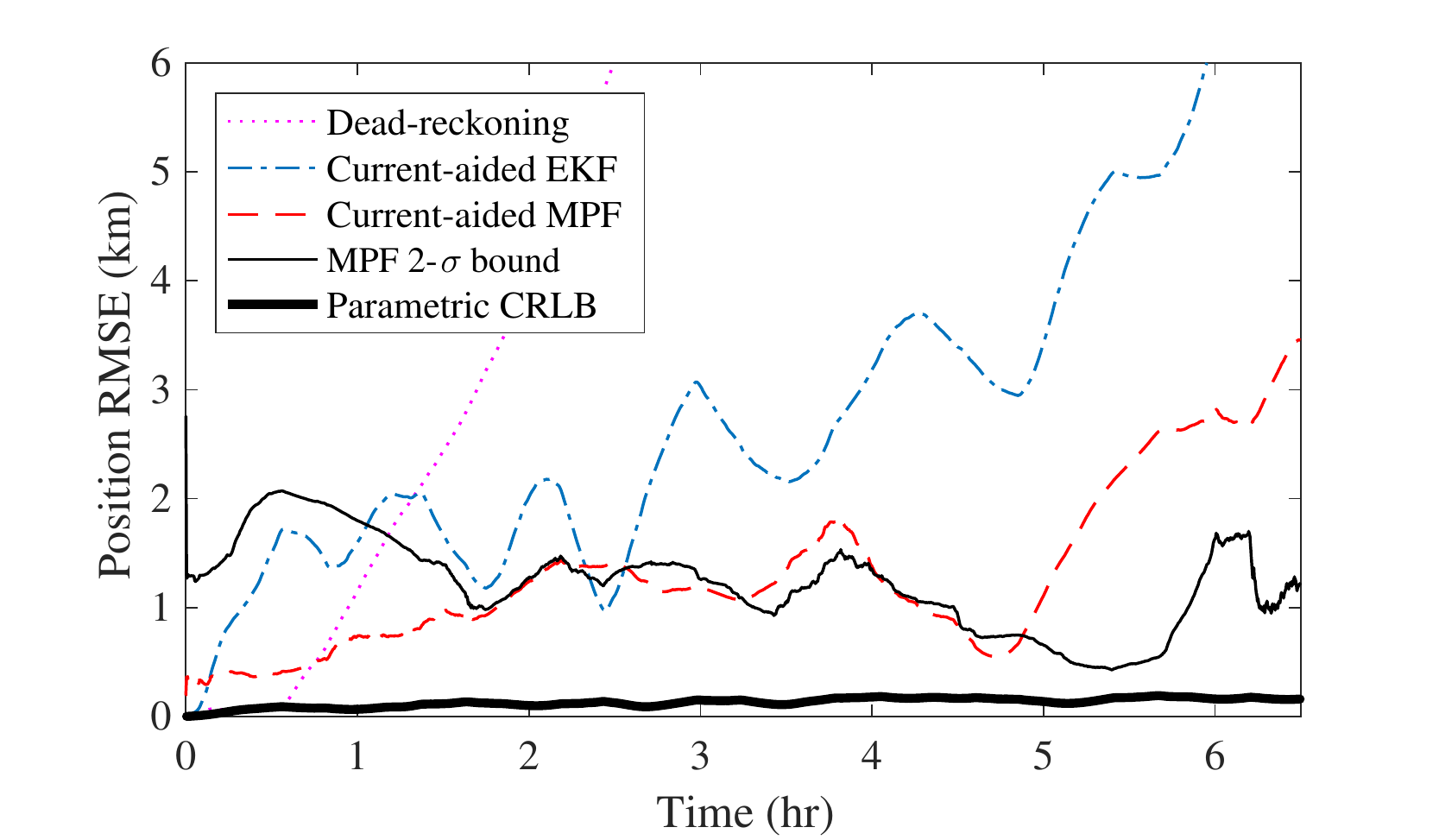}
\caption{Position RMSE and parametric CRLB for a vehicle navigating in a turbulent meandering jet flow field. Estimation results were averaged based on 50 Monte Carlo simulations with random particle initial locations and random KS turbulent flow components.}
\label{fig:MJpRMSE}
\end{figure}

\begin{figure}
\centering
\includegraphics[width = 0.7\linewidth]{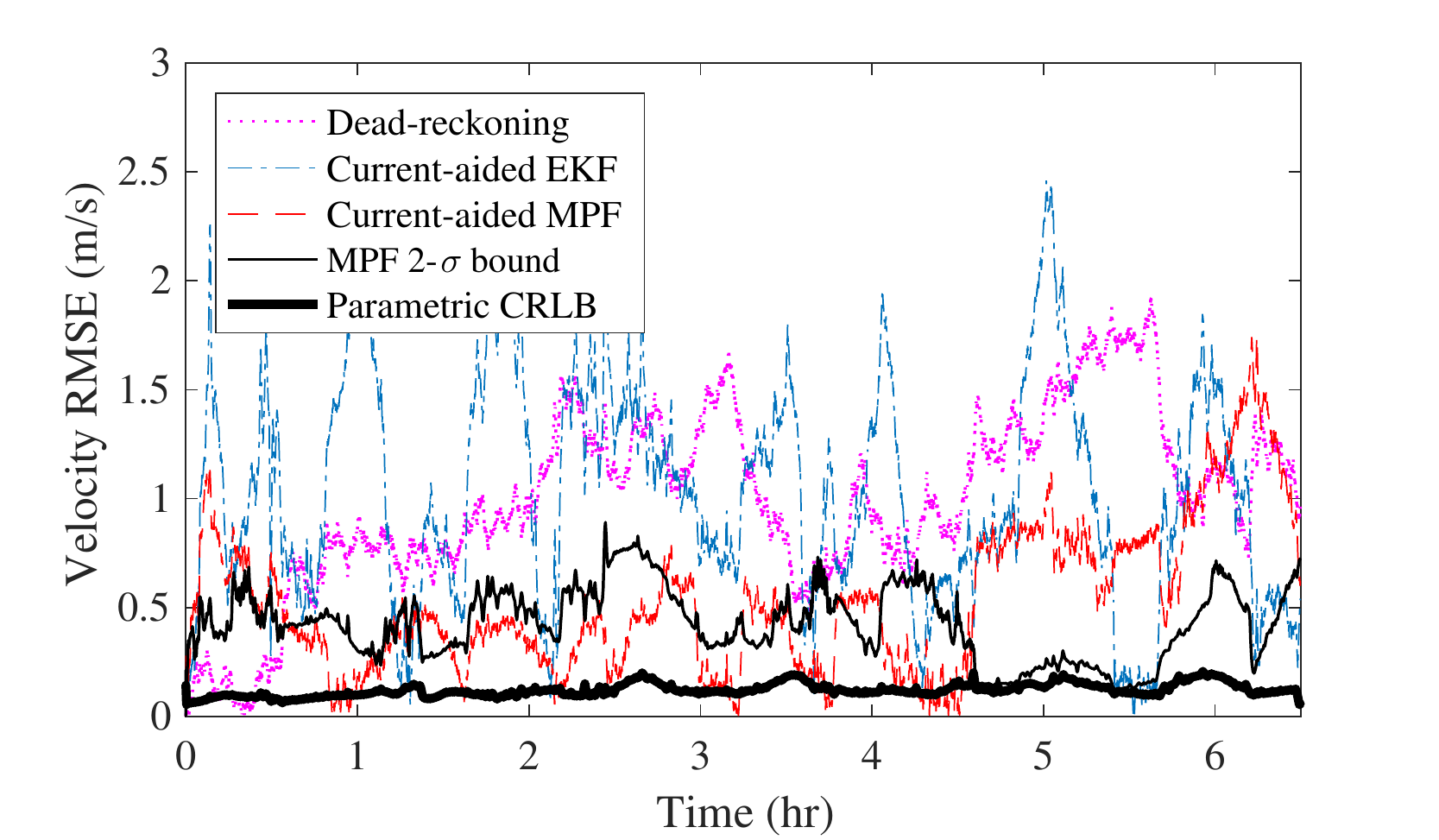}
\caption{Vehicle velocity RMSE and parametric CRLB for a vehicle navigating in a turbulent meandering jet flow field.}
\label{fig:MJvRMSE}
\end{figure}

The path estimates under difference schemes in one simulation with the turbulent meandering jet are shown in Fig.~\ref{fig:MJpath}. Although the current-aided MPF is still able to provide salient improvement in navigation performance compared to both dead-reckoning and current-aided EKF, the overall estimation performance degenerates compared to the case in the double-gyre flow field. This is more obvious in Fig.~\ref{fig:MJpRMSE} and Fig.~\ref{fig:MJvRMSE} where the RMSE of position and velocity estimates were evaluated based on 50 Monte Carlo runs. Generally, larger $2$--$\sigma$ bounds were achieved. The fact that the CRLB for this case seems to be smaller than the previous case with the double-gyre flow field is considered to be due to the low SNR in the turbulent meandering jet since the CRLB analysis did not include the turbulent effect.

Another key observation is the emergence of estimation divergence in both position and velocity estimates starting after approximately 5 hours. This is partially due to the low SNR and high repetition in similar flow patterns in the turbulent meandering jet flow field. These effects not only result in large belief bounds at the early stage of the estimation, they also lead to particle deprivation around the true vehicle location. Fortunately, real-world ocean currents often contain irregularities in the mean flow structure that may relieve these undesirable phenomena observed in the simulation case with turbulent meandering jet flow field. Other potential remedies include appropriately increasing the number of particles, dithering the particles intermittently, and using more sophisticated resampling strategies. This topic will be further investigated in more detail in a future study.

\section{Evaluation with Field Test Data}\label{sec:experiment}
\begin{figure}
\centering
\includegraphics[width = 0.65\linewidth]{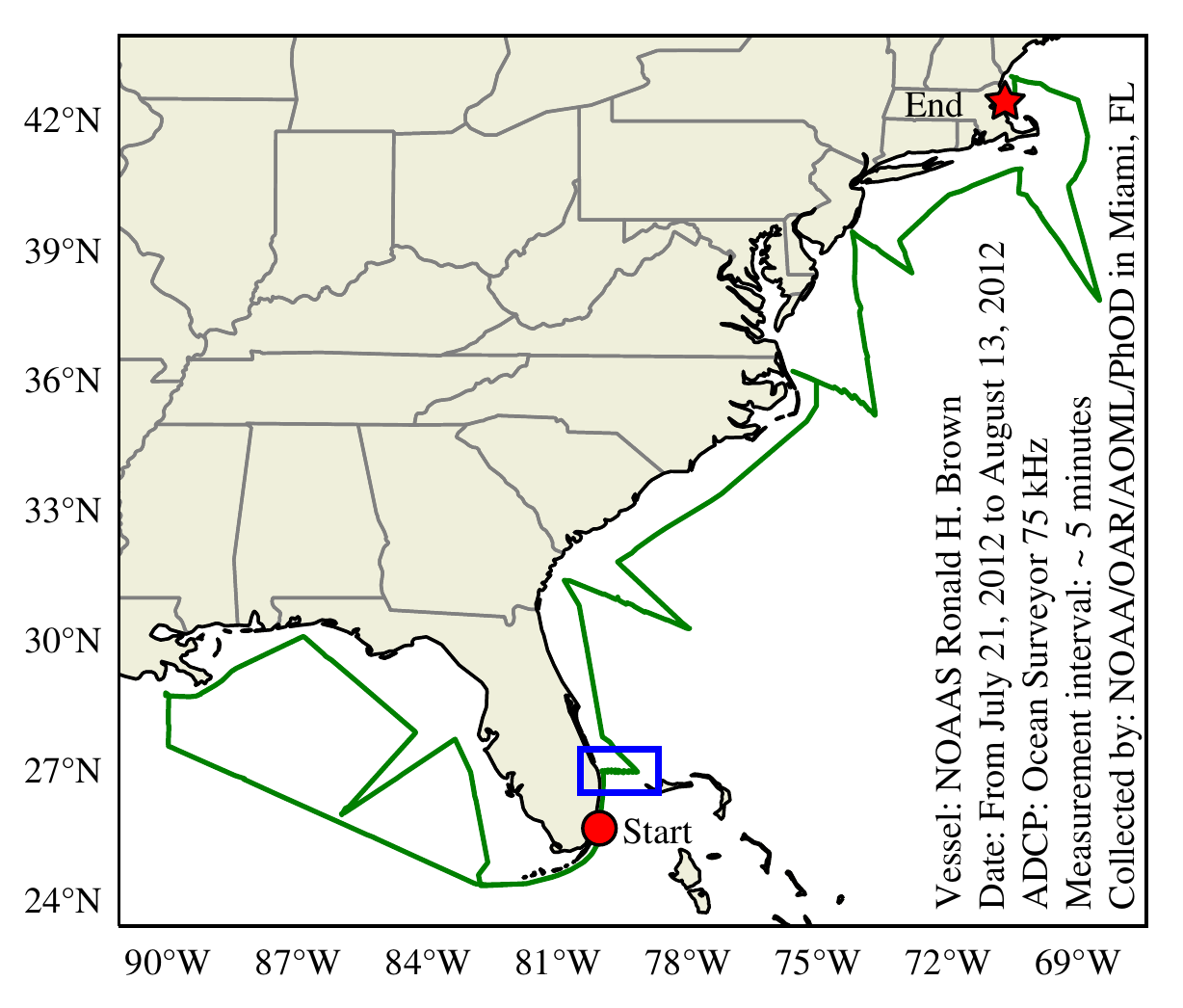}
\caption{Trajectory of the surface research vessel, NOAA Ship \textit{Ronald H.~Brown}, during survey project \textit{Gulf of Mexico and East Coast Carbon Cruise No.~2}.  The blue box indicates the segment used in this analysis.}
\label{fig:PathInMap}
\end{figure}

To further evaluate the current-aided navigation scheme under more realistic conditions, we constructed a simulated experiment based on field test data from survey project \textit{GOMECC-2 (Gulf of Mexico and East Coast Carbon Cruise No.~2)} conducted by NOAA/OAR/AOML/PhOD in Miami, FL from July 21 to August 13 in 2012.  During this project, a surface research vessel, NOAA Ship \textit{Ronald H.~Brown}, cruised along the coast of the Gulf of Mexico and the Atlantic coast in support of the coastal monitoring and research objectives of NOAA.  The survey vessel was equipped with a \textit{Teledyne/RD Instruments Ocean Surveyor 75 kHz} ADCP and the average interval between sequential measurements was approximately 5 minutes.  We applied the current-aided navigation scheme to this survey vessel to emulate an AUV performing mid-depth navigation.  The actual track of the vessel based on the recorded way points is shown in Fig.~\ref{fig:PathInMap}, where the blue box indicates the segment  (Segment-3) selected for the following analysis.  Data used for constructing the following experiment was obtained from the \textit{Joint Archive for Shipboard ADCP} hosted by NOAA and the University of Hawaii.  

\begin{figure}
\centering
\includegraphics[width = 0.6\linewidth]{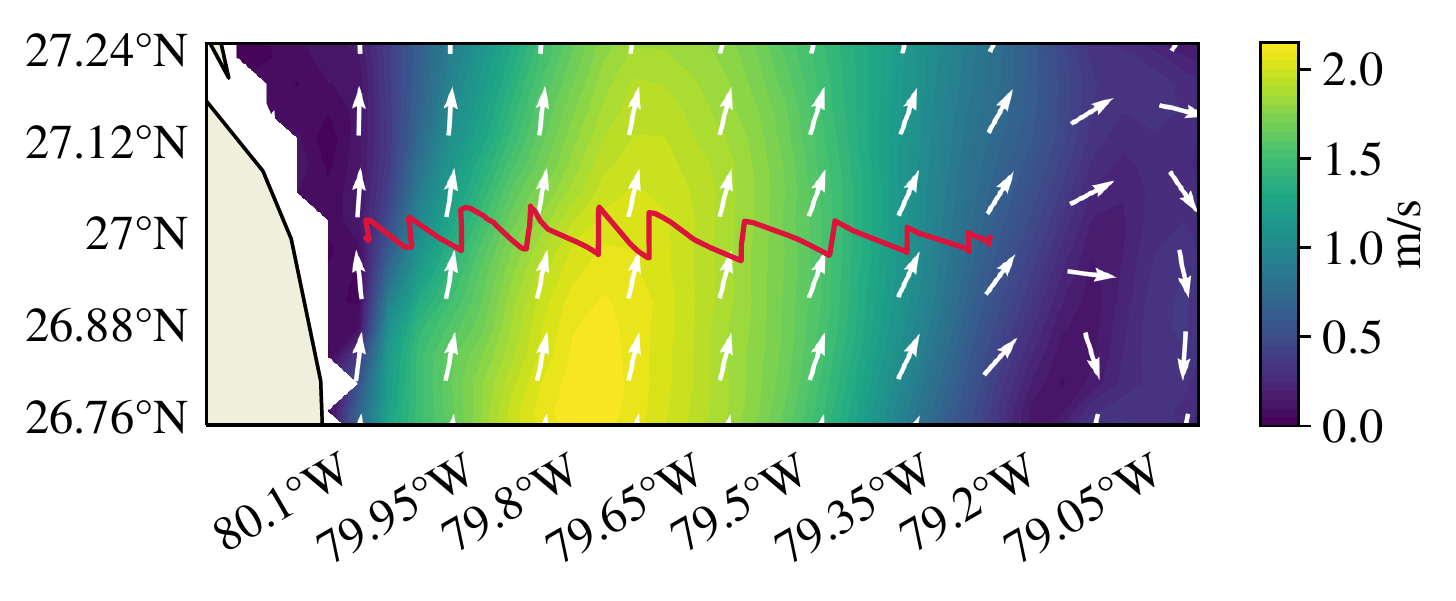}
\caption{A snapshot of the OGCM estimate of current velocity at depth $z = 10$ m of the selected region and a segment of the survey vessel track traversing the Gulf Stream.  Arrows indicate the flow direction.  OGCM data were obtained from the \textit{HYCOM + NCODA Gulf of Mexico 1/25$^\circ$ Analysis (GOMl0.04/expt_31.0)}.}
\label{fig:PathInFlow}
\end{figure}

\begin{figure}
\centering
\includegraphics[width = 0.6\linewidth]{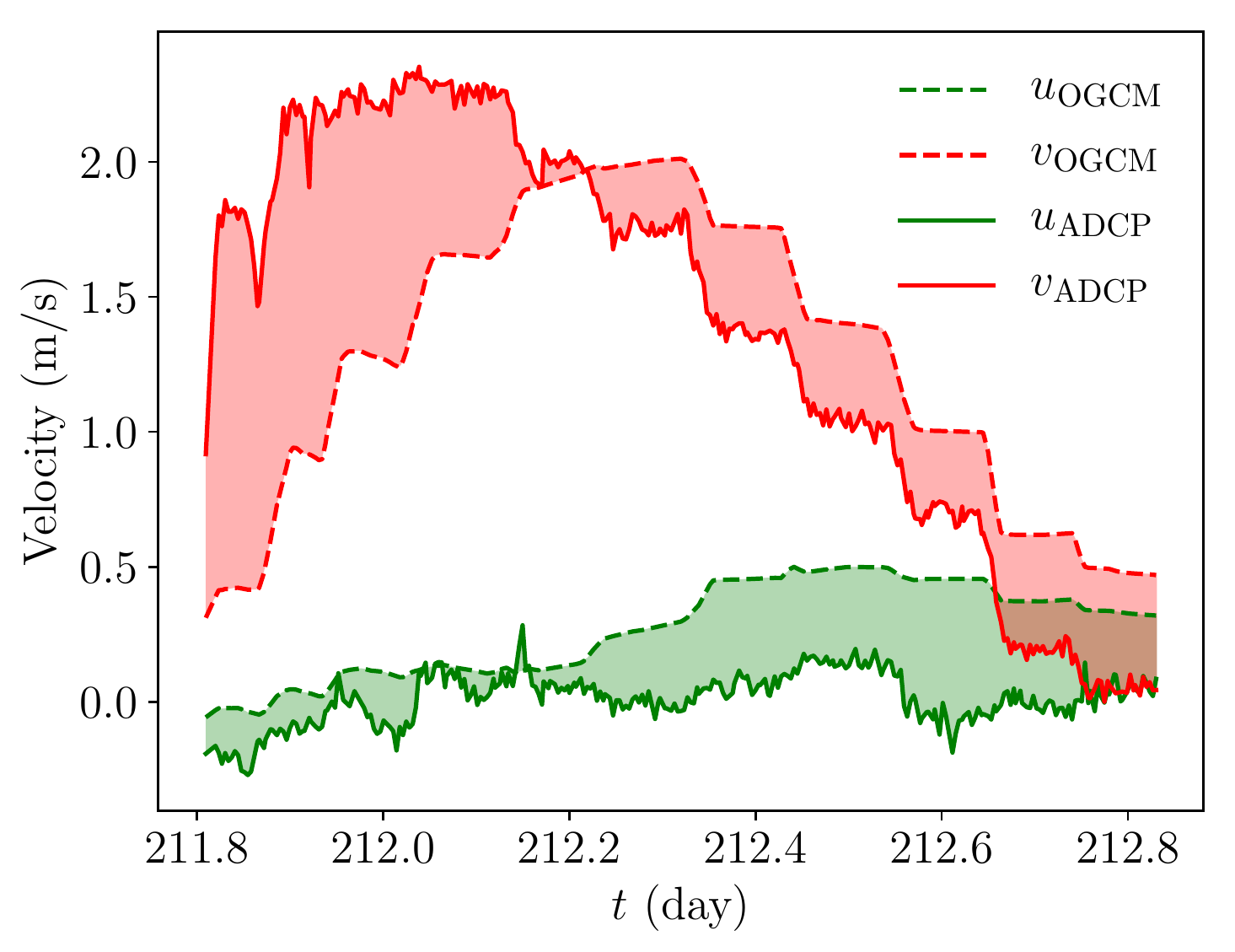}
\caption{Horizontal components of the current velocity along the trajectory of the survey vessel. Shaded areas indicate the estimation discrepancy due to flow phenomena unresolved by the OGCM.}
\label{fig:OGCMError}
\end{figure}

We accompanied the ADCP data from the GOMECC-2 project with the ocean current estimates produced by the \textit{HYCOM + NCODA Gulf of Mexico 1/25$^\circ$ Analysis (GOMl0.04/expt_31.0)} for the same period of time.  The publicly available Gulf of Mexico model has the spatial resolution of approximately 3.5~km at mid-latitudes and the temporal resolution of 1 hour.  Both the ADCP measurements and OGCM analysis at the depth of 10~m were selected for this experiment. Fig.~\ref{fig:PathInFlow} shows a snapshot of the OGCM estimate of the current velocity at $z = 10$ m in the selected region and the chosen segment of the survey vessel track.  The ship followed a zigzag trajectory while it traversed the Gulf Stream. Tri-linear interpolation was applied to obtain the current velocity estimated by the OGCM at any given location and instant of time.  Fig.~\ref{fig:OGCMError} compares the horizontal components of the current velocity between large-scale OGCM analysis and in-situ ADCP measurements along the track of the survey vessel. Shaded areas indicate the estimation discrepancy due to flow phenomena unresolved by the OGCM, which will be tracked by  $\hat{\boldsymbol{u}}^{\{n\}}_{c,k}$ in the system state vector.  

Considering the fact that the selected vessel track has a timespan of approximately 25 hours, a more accurate INS, \textit{VectorNav VN-110}, was selected for reference during sample generation.  Potential error effects due to sensor misalignment were taken into consideration by augmenting the bias instability characteristics of both the accelerometer and the gyroscope such that $\sigma_{b}^{\text{aug}} = \sigma_b + \sigma_b^+$.  It is worth mentioning that the ADCP measurement data acquired from the GOMECC-2 project had already been calibrated to represent the absolute current velocities.  Therefore, these measurements were first converted to relative current velocities represented in the vehicle's body frame using true vehicle velocity and heading, and were then corrupted with artificial noise components following the aforementioned method to generate ADCP samples for the simulated experiment.  Sensor characteristics used in generating the sample data are tabulated in Table~\ref{tb:INSparameterFieldTest}, where only the items with different values from the previous test cases are shown. 

\begin{table}
\centering
\caption{Parameters for sensor sample generation used in the simulation based on field test data. }
\begin{tabular}{l  l}
\toprule[0.75mm]
INS  & VectorNav \textit{VN-110/210}\\
\hline
Accel. bandwidth & 240 Hz \\
Accel. white noise SD ($\sigma_a$) & $0.04$ mg/$\sqrt{\text{Hz}}$ \\
Accel. in-run bias stability ($\sigma_{b_a}$) & 10 $\mu$g \\
Gyro. bandwidth & 240 Hz \\
Gyro. white noise SD ($\sigma_r$) & 3.24 ${}^\circ$/hr/$\sqrt{\text{Hz}}$ \\
Gyro. in-run bias stability ($\sigma_{b_r}$) & 1 ${}^\circ$/hr \\
\hdashline
Accel. misalignment ($\sigma_{b_a}^+$) & 0.03999 g \\
Gyro. misalignment ($\sigma_{b_r}^+$) & 59 ${}^\circ$/hr \\
\midrule[0.5mm]
ADCP & RDI \textit{75 kHz}\\
\hline
Measurement uncertainty ($\sigma_{\text{z}}$) & 0.005 m/s\\
Update rate used & $\sim$5 min \\
\bottomrule[0.75mm]
\end{tabular}
\label{tb:INSparameterFieldTest}
\end{table}

\begin{figure}
\centering
\includegraphics[width = 0.55\linewidth]{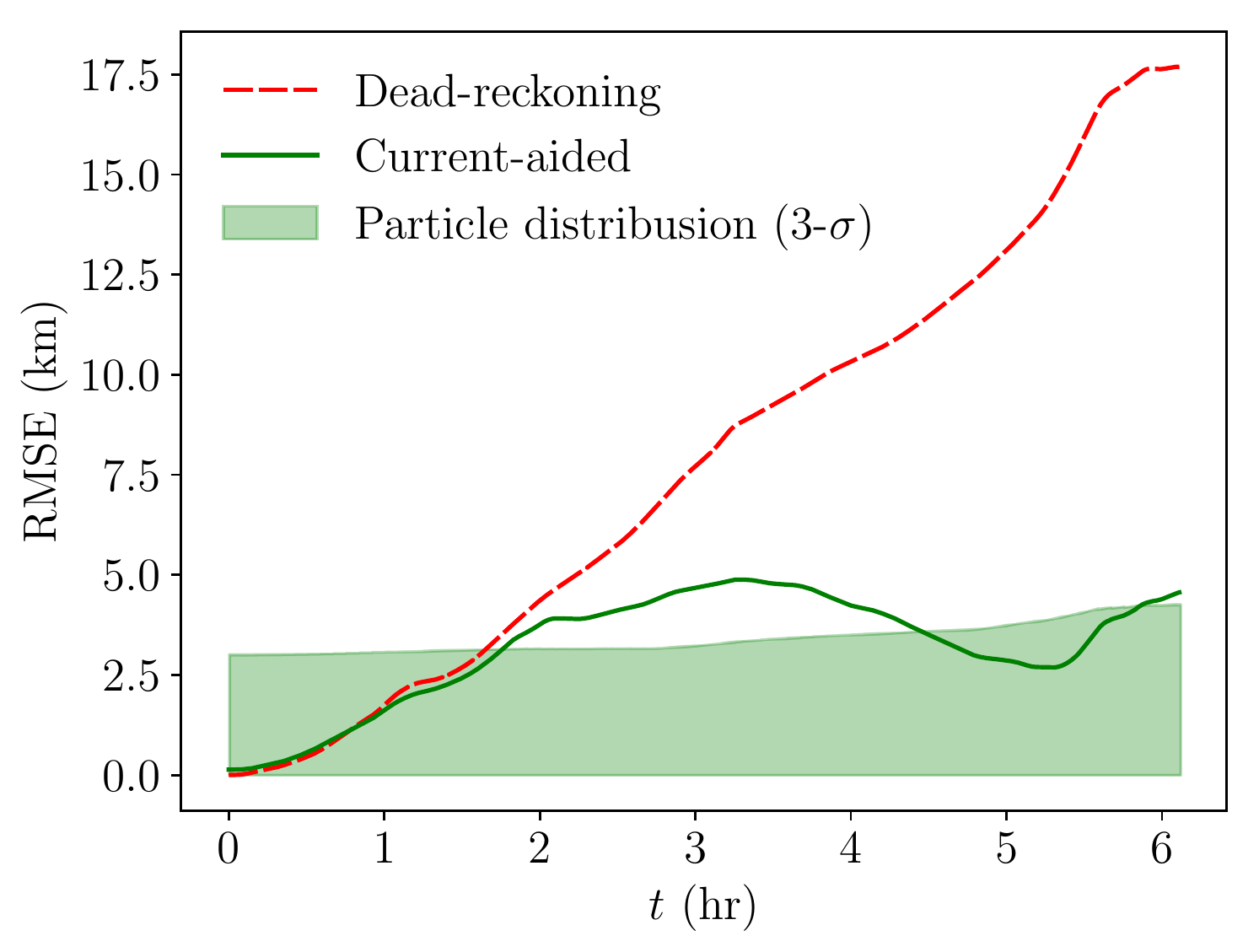}
\caption{Navigation error estimated by both dead-reckoning and the current-aided scheme. Standard deviation of the particle swarm from its weighted average is represented by the shaded area to show changes in particle distribution.  The current-aided navigation scheme resulted in positioning error within 8\% UDT, nearly 76\% reduction compared to the dead-reckoning performance with only INS.}
\label{fig:EstimationErrorShort}
\end{figure}

Performance of the current-aided navigation scheme for relatively short-term navigation ($\sim6$ hours), similar to the previous testing cases, was first evaluated.  A total number of 50 particles were initialized with EKF states:
\begin{align*}
&\boldsymbol{v}^{[i]}_0 \sim \mathcal{N}(\boldsymbol{v}^\text{true}_0,\, 10^{-6}I_{2\times1})\,\text{m/s}, \quad \psi^{[i]}_0 \sim \mathcal{N}(\psi^\text{true}_0,\, 10^{-8}),\\
&\boldsymbol{b}^{[i]}_{\boldsymbol{a},0} = \mathbf{0}_{2\times1}, \quad b^{[i]}_{r,0} = 0, \quad \boldsymbol{b}^{[i]}_{\boldsymbol{z},0} = \mathbf{0}_{2\times1}, \\
&\boldsymbol{u}^{[i]}_{c,0} \sim \mathcal{N}(\boldsymbol{u}^\text{true}_{c,0},\, 10^{-4}I_{2\times1})\,\text{m/s},
\end{align*}
and at random locations of $\boldsymbol{p}^{[i]}_0 \sim \mathcal{N}(\boldsymbol{p}^\text{true}_0,\, 10^6I_{2\times1})$~m in UTM coordinates of zone 17R.  Fig.~\ref{fig:EstimationErrorShort} shows the navigation uncertainty for 6 hours estimated by both dead-reckoning and the current-aided scheme. Standard deviation of the particle swarm from its weighted average is represented by the shaded area to show changes in particle distribution.  At an early stage of the mission, the state estimator was heavily weighted towards the dead-reckoning result due to strong confidence in the initial states of the vehicle and higher accuracy in the INS than the background flow reference.  The current-aided effect started to become evident after about 2 hours.  The proposed approach resulted in vehicle position with under approximately 8\% uncertainty per distance traveled (UDT), nearly 76\% reduction compared to the dead-reckoning performance with only the INS.  This performance is consistent with the result obtained from the previous simulations in the meandering jet flow field ($<$7.3\% UDT).  The performance degeneration compared to the artificial test case in the double-gyre flow field ($<$3\% UDT) was likely due to the lack of variation in background current velocities along the chosen segment of the vessel's trajectory.   

\begin{figure}
\centering
\includegraphics[width = 0.55\linewidth]{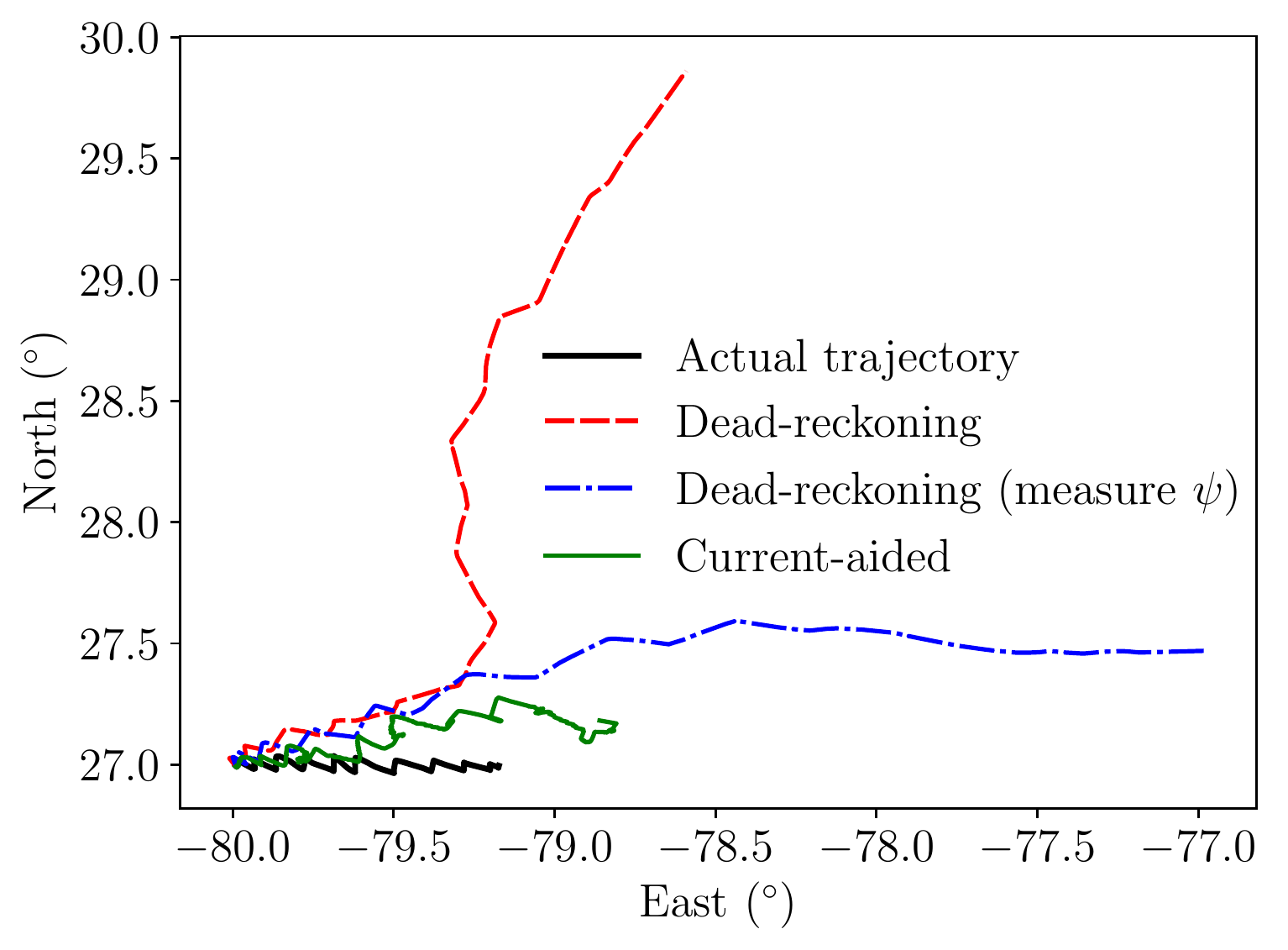}
\caption{Vehicle trajectories based on actual way points, dead-reckoning estimates using only noisy INS samples, dead-reckoning estimates with direct heading measurements, and estimates from the current-aided navigation scheme when provided direct heading measurements. }
\label{fig:TracksGeoTruePsi}
\end{figure}

\begin{figure}
\centering
\includegraphics[width = 0.55\linewidth]{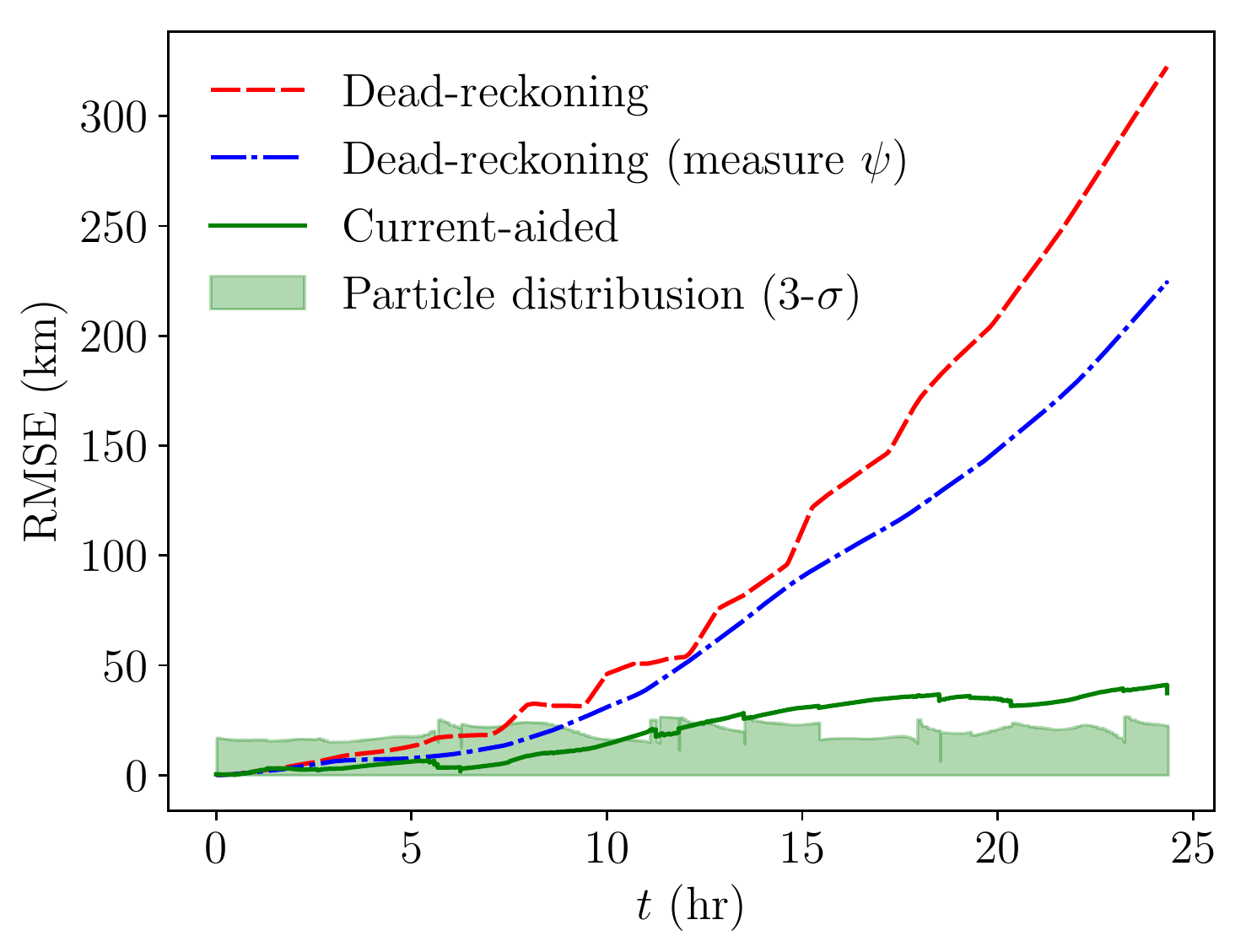}\\
\caption{Positioning errors and changes in particle distribution.  The vehicle's position was estimated based on the weighted average of all particles.  With independent heading measurements, the current-aided navigation scheme resulted in vehicle position estimates with under approximately 25\% UDT, approximately 83\% smaller than the performance of dead-reckoning with heading measurements. Intermittent increases in the particle distribution were the result of particle mutation, which leads to slight decreases in positioning error under the current-aided navigation scheme. }
\label{fig:EstimationErrorWithTruePsi}
\end{figure}

We then extended the mission time to approximately 24 hours by considering the entire segment of the vessel's track shown in Fig.~\ref{fig:PathInFlow}. For navigation missions with such a long timespan, it is valid to assume that the vehicle is also equipped with an accurate heading reference sensor such that $\psi_k$ can be independently measured.  To alleviate the ``particle deprivation" issue that potentially occurs during long-term state estimation, a ``mutation" step was introduced to dither the locations and the EKF states of the particles when the standard deviation of particle locations drops below a certain threshold.  Comparison between the estimated vehicle trajectories is shown in Fig.~\ref{fig:TracksGeoTruePsi}.  Fig.~\ref{fig:EstimationErrorWithTruePsi} shows the positioning errors and the changes in particle distribution.  When provided independent heading measurements, the current-aided navigation scheme resulted in vehicle position estimate with under approximately 25\% UDT, which is about 83\% smaller than the dead-reckoning performance with heading measurements.  Benefits of particle mutation can be noted as intermittent jumps in the particle distribution accompanied by slight decreases in positioning error. 

\begin{figure}
\centering
\includegraphics[width = 0.55\linewidth]{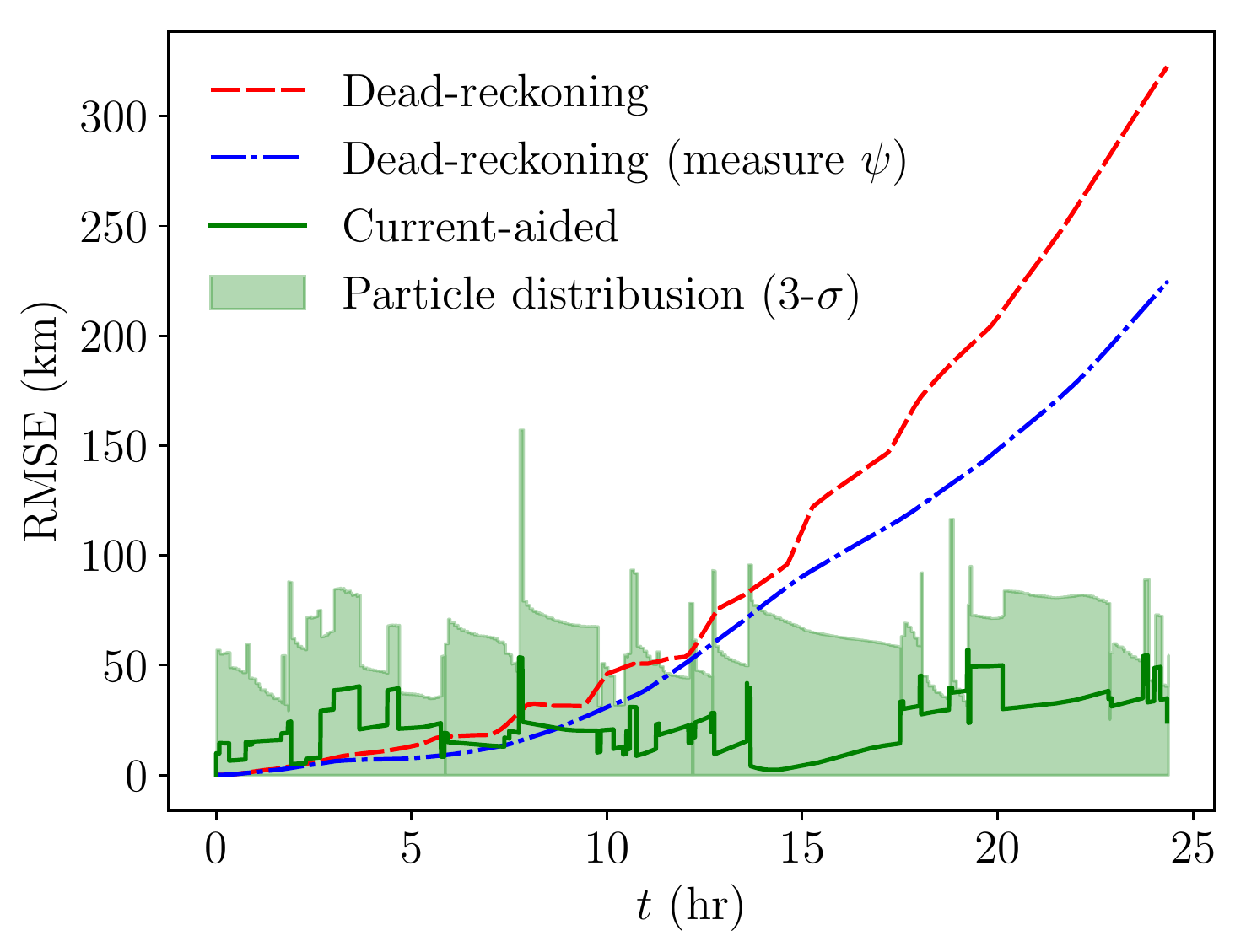}
\caption{Positioning errors and changes in the particle distribution when using a larger standard deviation for the particle distribution.  The vehicle's position was estimated based on the particle with the largest weight.  With independent heading measurements, the current-aided navigation scheme resulted in positioning estimates within approximately 16\% UDT, which is about 89\% smaller than the performance of dead-reckoning with heading measurements.}
\label{fig:EstimationErrorWithTruePsiBestParticle}
\end{figure}

So far, we have been focusing on maintaining satisfactory navigation accuracy throughout the entire mission.  This was targeted at long-term sampling missions where the positioning accuracy plays a key role in geo-referencing the collected environmental data along the vehicle's trajectory.  There are situations where terminal positioning accuracy is a higher priority.  To accommodate these demands, we investigated the performance of our proposed approach using particle distributions with a larger standard deviation.  This was motivated by the observation of Fig.~\ref{fig:EstimationErrorWithTruePsi} that the positioning error exceeds the $3$--$\sigma$ bound of the particle distribution after about 10 hours.  In this test case, the vehicle's position was estimated based on the particle with the largest weight.  The particles were initialized at locations of $\boldsymbol{p}^{[i]}_0 \sim \mathcal{N}(\boldsymbol{p}^\text{true}_0,\, 10^8I_{2\times1})$~m.  When the standard deviation of the particle swarm drops below 10~km, we allowed the particles to ``mutate" to $\boldsymbol{p}^{[i],\,\text{m}}_k \sim \mathcal{N}(\boldsymbol{p}^{[i]}_k,\, 10^8I_{2\times1})$~m.  As can be observed in Fig.~\ref{fig:EstimationErrorWithTruePsiBestParticle}, the terminal positioning performance achieved by the current-aided scheme is around 16\% UDT, which is approximately 89\% smaller than the result of dead-reckoning with independent heading measurements.  This was achieved at the cost of higher positioning uncertainty at an early stage of the mission.  

It is worth noting that, although the ocean flow estimation error can be huge (Fig.~\ref{fig:OGCMError}), the current-aided navigation scheme has shown to still be able to provide performance improvement over long-term missions. There are three major factors that enable such robustness of the ocean-aided navigation scheme over the erroneous ocean model. Firstly, when performing current aiding, the ocean flow velocity estimate is utilized in combination with the turbulence component estimate ($\hat{\boldsymbol{u}}^{\{n\},-}_{c,k+1}$ in \eqref{eq:dynamicsEnd}), which is included in the system state vector to track the flow estimation error by the ocean model. Especially when  $\boldsymbol{u}^{\{n\}}_{c}$ is initialized with the true value, the effect of ocean model estimation inaccuracy towards the overall navigation performance is reduced. Long-term effect of ocean model estimation inaccuracy, however, can become significant, which can be observed from the comparison between the navigation performance for the first 6 hours (8\% UDT) and the entire 24 hours (25\% UDT).  Secondly, information fusion between INS-based inference and current-aiding feedback is handled through the EKFs, which weigh the state estimator's ``confidence" between dead-reckoning and the ocean model by proper selection of values for $Q$ in \eqref{eq:P_update} and $R$ in \eqref{eq:innovationCov}. This prevents the state estimator from becoming overconfident in ocean model estimate before the dead-reckoning uncertainty increases to the level of ocean model inaccuracy. Lastly, by tracking the ocean flow estimation error, the current-aided navigation scheme embeds an effect similar to water-tracking since $\boldsymbol{u}^{\{n\}}_{c}$ is modeled as a slow-varying process coupled with the vehicle's motion. This further improves the estimator's robustness over high-frequency INS noise components. It is among our current interests to systematically identify the relation between state estimation error and ocean model uncertainty with a more generalized formulation, the details of which will be discussed in a future publication.

We should point out that, although designed based on field test results, the experiment conditions considered in this performance analysis bear some differences from an actual mid-depth navigation mission.  Firstly, since project GOMECC-2 was conducted by a surface research vessel, the ADCP measurements and the OGCM estimates used in this test case were of the ocean
current velocities in the uppermost layer or the epipelagic zone.  OGCMs may have superior accuracy in estimating surface currents than mid-depth currents due to relatively more frequent (both spatially and temporally) data assimilation thanks to remote sensing and several forms of in-situ measurements. The current-aided navigation approach respects the accuracy issue of ocean models and tracks the error in current velocity estimation online along with other system states. As this testing case suggests, even with large errors in current velocity estimation, the benefit of current-aiding to an INS is still rather significant for long-term navigation.  Given the recent progress in ocean modeling and technologies for in-situ oceanographic data collection~\cite{TestorP:10a,WynnRB:14a,BuckleyMW:16a}, it is reasonable to believe that the accuracy of ocean current estimate will further improve over time.  As a result, larger improvements in navigation performance can be expected using the proposed approach as OGCMs further develop in the near future. 

Secondly, OGCM data used in this analysis were the nowcast results after data assimilation for the period of GOMECC-2 project.  Forecast results, however, may have inferior accuracy that decreases as the forecast lead time increases.  It should be noted that even after data assimilation, as  indicated by the nowcast results used in this analysis, discrepancy between the estimated current velocity and in-situ measurements, as large as 1.5~m/s in magnitude in this case, may still be nontrivial due to model-unresolved flow phenomena.  Tracking this discrepancy as a system state allows the current-aided navigation scheme to benefit from useful information in large-scale circulation with certain robustness.  Recent studies on the validity of the Global Ocean Forecast System has reported current forecast RMSE under 0.35 m/s for the Gulf Stream~\cite{MetzgerEJ:17a}, which is well below the uncertainty level under which our approach has been evaluated.  More rigorous, theoretical analysis of the accuracy of ocean model predictions is an active research topic in oceanography.  Interested readers are referred to a recent study by Wei et al.~\cite{WeiM:16a} and the references therein for more details. 

Meanwhile, the current-aided navigation scheme can be further extended to potentially improve its performance even with the state-of-the-art OGCM products.  One of our on-going efforts is the extension to a multi-vehicle scenario where information exchange among a team of AUVs allows each vehicle to reference the current velocity maps on a larger scale~\cite{Song:17c}.  Local flow velocity estimates by a navigating vehicle can also be utilized for on-board, small-scale data assimilation in order to maintain or even improve the accuracy of the preloaded current velocity maps.  We will explore these extensions in detail in a future publication.

\section{Concluding Remarks}\label{sec:conclusion}
In this paper, we proposed the concept of an aided navigation system utilizing the dynamics of a continuous flow field. An ocean current aided inertial navigation system was presented for an autonomous underwater vehicle (AUV) in long-term, mid-depth navigation where only dead-reckoning is available. It aims at improving the inertial navigation performance through estimating ambient flow velocities and referencing preloaded background flow velocity maps predicted by ocean general circulation models. A recursive Bayesian structure was formulated for the current-aided state estimator.  The vehicle's states, including sensor bias and unmodeled, small-scale turbulence, were tracked based on the knowledge of the vehicle's motion propagation model, the relative flow velocity measurement model, and the preloaded velocity maps of large-scale ocean currents. We implemented this Bayesian structure as a semi-parametric state estimator based on a marginalized sequential Monte Carlo framework, i.e. a marginalized particle filter (MPF), to circumvent linearizing a non-analytical measurement model based on digital flow maps. 

The performance of the current-aided MPF was first carefully evaluated with noisy sensor samples generated based on the characteristics of an automotive-grade INS, and two turbulent flow fields that were considered to resemble the behavior of ocean currents: a double-gyre flow field and a meandering jet flow field. Unmodeled, small-scale flow perturbations with Kolmogorov energy spectrum were added to the mean flow component. The proposed navigation scheme has been demonstrated to provide significant performance improvement over the dead-reckoning method in both cases. With a large mean-flow gradient along the path of the vehicle, the current-aided MPF has been shown to achieve near optimal filtering performance asymptotically when compared to the parametric Cram\'{e}r-Rao lower bound. We found that a larger spatial variation in flow velocities tends to result in a smaller variance in localization error. This analysis can also be used for predicting the theoretical navigation performance given a series of general circulation maps, a prescribed vehicle path, and the accuracy of sensor measurements. Through a more realistic, simulated experiment based on field test data and actual ocean modeling results, the feasibility and performance of our proposed method were further demonstrated.

To the best of our knowledge, this is the first effort to investigate and demonstrate the feasibility and limitations of an underwater navigation system using ocean current forecast as localization reference. This approach can also be implemented in conjunction with existing current-aided navigation schemes aiming at more accurate vehicle velocity tracking (e.g.~\cite{HegrenaesO:09a} or~\cite{Medagoda:16a}) to further improve the navigation performance of an AUV. In long-distance, mid-depth AUV missions where neither frequent surfacing nor constant bottom-tracking is available, the proposed current-aided navigation concept can be applied to improve the inertial navigation performance. With the progress of high-definition ocean modeling and forecasting, the proposed approach may be adopted as one of the primary navigation schemes for underwater vehicles.

\section*{Acknowledgment}
This material is based on work partially supported by the Office of Naval Research under grant award number N00014-16-1-2083 and the National Science Foundation under grant award number NRI-1638034.  The authors would like to thank the GOMECC-2 project team for publishing their survey data, and to thank Dr.~Ryan Smith from the University of Hawaii and Mr.~Patrick Caldwell from NOAA for providing access to the data.  The authors would also like to thank the anonymous reviewers for their valuable comments that greatly helped improve the quality of this manuscript.

\bibliographystyle{ieeetran}
\bibliography{RefA2,RefSong}

\end{document}